\documentclass[11pt]{article}

\usepackage{amssymb}
\usepackage{amsmath}
\usepackage{pdfpages}
\usepackage{caption}
\usepackage[letterpaper,top=2cm,bottom=2cm,left=3cm,right=3cm,marginparwidth=1.75cm]{geometry}

\usepackage{amsmath}
\usepackage{graphicx}
\usepackage[colorlinks=true, linkcolor=black, citecolor=blue]{hyperref}
\usepackage{color}
\usepackage{xcolor}
\usepackage{multirow}
\usepackage{booktabs}
\usepackage{xspace}
\usepackage{setspace}
\usepackage[superscript]{cite}
\usepackage{float}

\usepackage{graphicx} 
\usepackage{caption}  

\usepackage{enumitem}
\newcounter{suppfigure}

\usepackage{titletoc}

\newcommand{\mycomment}[1]{}
\newcommand{\method}{\text{CTO}\xspace}

\makeatletter
\renewcommand{\maketitle}{\bgroup\setlength{\parindent}{0pt}
\begin{flushleft}
  \textbf{\@title}

  \@author
\end{flushleft}\egroup
}
\makeatother

\usepackage{pifont}
\newcommand{\cmark}{{\color{teal}\ding{51}}}
\newcommand{\xmark}{{\color{red}\ding{55}}}
\usepackage{bm}

\title{Automatically Labeling Clinical Trial Outcomes: A Large-Scale Benchmark for Drug Development}

\author{Chufan Gao$^{*}$, Jathurshan Pradeepkumar$^{*}$, Trisha Das$^{*}$, Shivashankar Thati and \\ Jimeng Sun\\
Department of Computer Science, University of Illinois Urbana-Champaign, Urbana, IL, USA \\
$^*$ Equal contribution \\
Corresponding author: \texttt{jimeng@illinois.edu}}

\begin{document}

\maketitle
\begin{abstract}
\noindent\textbf{Background} The high cost and complexity of drug discovery and development make clinical trial outcomes critical for regulatory approval and patient care. However, the limited availability of large-scale, high-quality clinical trial outcome data hinders advancements in predictive modeling and evidence-based decision-making.
\\\\
\noindent\textbf{Methods} We introduce the Clinical Trial Outcome (CTO) benchmark, a fully reproducible, large-scale repository covering approximately 125,000 drug and biologics trials. CTO integrates large language model (LLM) interpretations of trial publications, phase progression tracking, sentiment analysis from news sources, stock price movements of trial sponsors, and additional trial-related metrics. To enhance outcome label quality and reliability, we manually annotated a dataset of clinical trials conducted between 2020 and 2024.
\\\\
\noindent\textbf{Results} The CTO benchmark demonstrates high agreement with expert annotations, achieving an F1 score of 94 for Phase 3 trials and 91 across all phases. Additionally, benchmarking standard machine learning models on our manually annotated dataset reveals distribution shifts in recent trials, emphasizing the need for continuous updates in trial outcome labeling.
\\\\
\noindent\textbf{Conclusions} Our analysis of CTO's performance on recent clinical trials highlights the ongoing need for high-quality, up-to-date trial outcome labels. We publicly release the CTO knowledge base and annotated labels at \url{https://chufangao.github.io/CTOD}, with regular updates to support research on clinical trial outcomes and drive data-driven advancements in drug development.
\end{abstract}

\newpage
\section*{Introduction}
Clinical trials are an indispensable yet high-risk stage in drug development, characterized by enormous costs and lengthy durations with only a small fraction of trials ultimately succeeding. Many trials fail due to issues of drug inefficacy, safety concerns, or challenges in patient recruitment. In 2022, drug discovery and development spending reached 244 billion dollars globally \cite{statista-pharma-rnd-expenditure}. Among these, the clinical trial market reached \$44.3 billion in 2020 and is expected to grow to \$69.3 billion by 2028 \cite{ltd2022}. 
Low efficacy, safety issues, and poor trial protocol design can lead to trial failures \cite{friedman2015fundamentals, sinha2023review, lu2023machine}. Eroom's Law\footnote{reverse of ``Moore’s Law''} shows that the number of new FDA-approved drugs per billion US dollars of R\&D spending has halved approximately every nine years since 1950, even with inflation adjustment~\cite{scannell2012diagnosing}. 
Despite this decline in approval efficiency, the volume of drug and biologics trials continues to grow each year. However, obtaining high-quality, human-labeled trial data remains expensive and largely inaccessible due to the limited public availability of detailed trial description and their outcomes. Addressing these challenges, collecting diverse and comprehensive clinical trial datasets and using AI models to automatically assess trial outcomes could help establish a high-quality benchmark for clinical trials, ultimately enhancing the efficiency of drug development.

The FDA does not release the clinical trial ID (NCTID) in documents of approved drug applications.
Other public data sources, such as the \url{ClinicalTrials.gov} database with more than 500,000 historical trials \cite{ross2009trial,zarin2011clinicaltrials,zarin2016trial}, provide vital information for identifying trial recruitment and design information but does not provide comprehensive trial outcome directly\footnote{The status is classified into basic categories: recruiting, completed, and terminated, but it does not clearly indicate whether the trial achieved the desired endpoints.}. 
The absence of a centralized, easily accessible knowledge base that consolidates clinical trial outcomes, drug approvals, and trial phase transitions poses a significant challenge for researchers and drug developers~\cite{chen2024trialbench}.
This fragmented information landscape hinders the development of accurate predictive models and can slow down the drug discovery and development process. 
Furthermore, the number of clinical trials conducted each year continues to rise, with their data dynamically evolving under the influence of various external factors, such as technological innovations and global events like the COVID-19 pandemic. This ever-changing landscape underscores the need for updated trial outcome labels to develop robust and high-performing clinical trial outcome prediction models.

\textbf{Trial Outcome Definition:} Clinical trial outcomes are multifaceted and have diverse implications. These outcomes can involve meeting the primary endpoint as defined in the study, advancing to the next phase of the trial, obtaining regulatory approval, impacting the financial outcome for the sponsor (either positively or negatively), and influencing patient outcomes such as adverse events and trial dropouts.
Our paper follows the previous conventions \cite{fu2022hint,wang2023spot,friedman2015fundamentals,lu2024uncertainty, lo2019machine, aliper2023prediction} and defines the trial outcome as a binary indicator, showing whether the trial achieves its primary endpoints and can progress to the next stage of drug development. For example, for Phase 1 and 2 trials, success may mean moving to the next phase, such as from Phase 1 to Phase 2, and from Phase 2 to Phase 3. In Phase 3, success is measured by regulatory approval. 

Recent efforts have been made to predict specific aspects of clinical trial outcomes. These include employing electroencephalographic (EEG) measures to predict the effects of antidepressant treatments \cite{rajpurkar2020evaluation}, optimizing drug toxicity predictions based on drug and target properties \cite{hong2020predicting}, and using phase II trial results to anticipate phase III trial outcomes \cite{qi2019predicting}. Additionally, there is a growing interest in developing comprehensive methods for predicting trial outcomes. For example, predicting drug approvals for 15 disease groups by analyzing drug and clinical trial features using classical machine learning techniques \cite{lo2019machine}, using multimodal drug structure and text information to predict outcomes based on a supervised set of data \cite{fu2022hint,wang2023spot}, 
an algorithm that computes the probability of technical success via asking experts a standardized questionnaire \cite{willigers2023algorithmic}, and multimodal trial outcome prediction via omics, text, clinical trial design, and small molecule properties \cite{aliper2023prediction}.

Despite these efforts, several limitations still impede the utility of existing trial outcome prediction models. Namely--the lack of transparency in the clinical trial labeling process. Fu et al.'s TOP \cite{fu2022hint} is one example of a publicly available expert-curated clinical trial dataset. However, it is also quite limited because it only contains around 12,000 human-labeled interventional small-molecule drug trials, most of which concluded prior to 2020. To date, we are unaware of any other large-scale, publicly available, open-source (fully reproducible) effort to compute trial outcome labels.

To address these limitations, we propose CTO, the first large-scale, publicly available, open-source, and fully reproducible dataset of clinical trial outcomes. We compiled a comprehensive trial knowledge base by integrating diverse data sources, including trial-related publications, news headlines, stock prices of trial sponsors, and other relevant trial metrics. From this knowledge base, we derive weakly supervised trial outcome labels using the CTO's automated labeling framework, which aggregates labels from trial phase linkages, LLM interpretations on publications, sentiment analysis of news headlines, p-values, and trial status, among other indicators. We validate CTO’s labeling accuracy against previously published trial outcomes, achieving an F1 score of 91 across all trial phases, closely aligned with human-annotated clinical trial outcomes. Additionally, aided by the CTO's framework, we manually curated a set of approximately 10,000 recent trials completed between 2020 and 2024. To ensure reproducibility and keep pace with rapidly evolving trial data, we will share regular updates of our trial knowledge base and labels generated by CTO’s automated labeling framework. 

\begin{figure}[t!]
    \centering
    \includegraphics[width=1\linewidth]{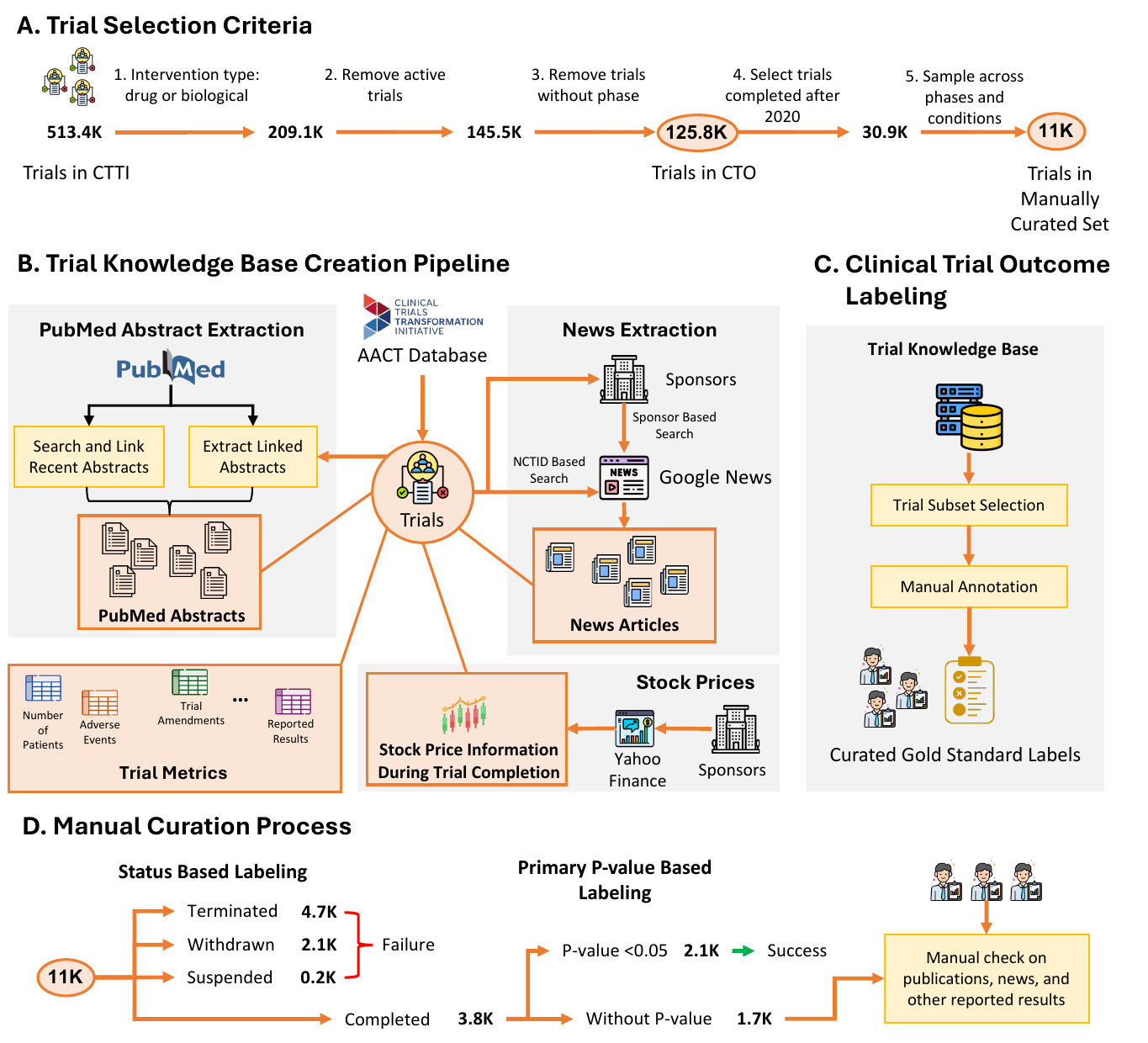}
    \caption{Overview of our CTO benchmark pipeline. (A) Trial selection criteria used to curate a targeted dataset from the CTTI database, narrowing from an initial 513.4K trials to 125.8K trials in CTO to a final manually curated set of 11K trials. (B) Trial knowledge base creation pipeline, including extraction of linked PubMed abstracts, recent abstract searches, trial metrics, news articles, and stock prices. (C) Clinical trial outcome labeling process, integrating the CTO automated labeling framework assisted manual labeling to generate curated gold standard labels. (D) The manual curation process, our rule-based annotation criteria, followed by manual checking on challenging clinical trials.}
    \label{fig:ctod1}
\end{figure}

\section*{Methods}

\subsection*{Trial Selection Process}
Our study focused primarily on clinical trials involving drugs and biologics. Starting with a dataset of approximately 513,485 clinical trials from the Clinical Trial Transformation Initiative (CTTI) database~\cite{ctti}, we applied a series of selection criteria to create a refined subset as illustrated in Figure~\ref{fig:ctod1}A. First, we selected only trials with interventions categorized as drugs or biologics, reducing the dataset to 209,124 trials. Next, we excluded active trials and those still recruiting to focus exclusively on completed studies, resulting in 145,464 trials. Subsequently, trials without phase information were removed, further narrowing the dataset to 125,840 trials, representing the number of trials labeled using our CTO's automated labeling framework. To construct a manually curated test set, we then selected trials completed after 2020, yielding a subset of approximately 30,900 recent trials. Importantly, we ensured this set did not overlap with the TOP benchmark\cite{fu2022hint}. Finally, we sampled different trial phases and conditions, producing a manually curated set of 11,012 trials.

\subsection*{Trial Knowledge Base Creation}
Based on the CTTI data, we generate a trial knowledge base by aggregating and linking various data sources to the trials, as shown in Figure~\ref{fig:ctod1}B. Our pipeline enables regular updates of the knowledge base given the updated CTTI data on a monthly frequency. The trial knowledge base extraction pipeline consists of four components: 1) PubMed abstract extraction, 2) News extractions, 3) Trial metrics, and 4) Stock price information extraction. 

\noindent\textbf{Pubmed abstract extraction:} PubMed abstracts have been automatically linked to trials by the Clinical Trials Transformation Initiative (CTTI)\cite{califf2012characteristics,tasneem2012database,anderson2015compliance} as well as other efforts \cite{huser2013linking}. Despite these sources, we found that many recent abstracts have not linked to their corresponding trials. To alleviate the linking issues, we perform the following procedure. First, we extracted all linked PubMed abstracts for each trial through the National Center for Biotechnology Information (NCBI) API \footnote{https://ncbi.nlm.nih.gov/}, and the statistics of the linked abstracts are given in the Supplementary. These abstracts can be categorized into 1) Background, 2) Derived, and 3) Results. Since we are interested in clinical trial outcomes, we only utilized abstracts in the Derived and Results categories. To enrich the knowledge base further, we performed regular searches on NCBI using NCT IDs to find newly published abstracts, linking any relevant ones to their associated trials. The initial count of drug or biologic trials with extracted PubMed abstracts linked from CTTI data was 49,277. Our additional search and extraction process linked 10,232 new trials with their recent PubMed abstracts.
Ultimately, our knowledge base comprises 41,228 Results-type abstracts, 65,327 Derived-type abstracts, and 20,512 newly linked abstracts.

\noindent\textbf{News extractions:} In our initial analysis, we identified news sources as a valuable resource for determining the outcome of clinical trials, especially since most trials in the CTTI dataset lack reported results. We extracted news articles related to completed drug and biologics trials by web scraping and linked them to the trials in our knowledge base. We sent requests to Google News for all completed drug and biologic trials (excluding phase 4), which accounted for around 85K trials. Due to rate limitations, we used an external API (SerpAPI) to streamline this process. Our findings revealed that most trials had no corresponding news coverage; only 3,688 were associated with news articles.

\noindent\textbf{Trial metrics:}
We extracted trial metrics from CTTI\cite{ctti} and mining clinicaltrials.gov. Specifically, we extracted nine numerical metrics--the number of patients, sponsors, sites, amendments, patient drops, deaths, serious adverse events, all adverse events, and the number of days the trial has been last updated after completion.
We also extracted three binary metrics--whether results were reported, the completion/terminated status of the trial, and finally, whether the trial had a significant p-value.
Detailed descriptions of each are in the supplementary.

\noindent\textbf{Stock price information extraction:} The stock price of a pharmaceutical or biotech company often reflects market expectations. If investors expect positive results, the stock may rise in anticipation of the trial's completion. Conversely, if expectations are low or if previous trials have been unsuccessful, the stock may not perform as well. We utilized Yahoo finance\footnote{\url{https://pypi.org/project/yfinance/}} to collect historical stock data for companies with publicly available tickers for completed trials. Details on the calculations involving stock price movements and their linkage to specific trials are provided in the supplementary.

\subsection*{CTO Automated Labeling Framework}

We developed a clinical trial outcome labeling pipeline that first assigns rule-based labels. In this paper, labels or labeling refer to the machine learning concept related to the ground-truth prediction target, unrelated to drug labels.
First, trials with statuses indicating {`terminated', `withdrawn', `suspended', `no longer available'\} were labeled as failures. 
Trials labeled as `approved for marketing' were marked as successes. 
For trials with `completed' as status, we examined the reported results: if any reported p-value for a primary endpoint was $<$0.05, the trial was labeled as a success. Unlabeled trials are then automatically labeled.
The main idea behind our automated labeling framework is to combine multiple sources of labels with varying levels of accuracy. 
We call these sources of labels ``weakly supervised labeling functions'' or LFs for short\cite{ratner2016data}.
Following this, we apply a label model that leverages our sources to refine labels and produce denoised pseudo-labels. Each LF, such as detecting efficacy based on p-value thresholds, assigns labels or abstains when uncertain. 
Certain metrics, such as adverse events, are also subject to phase-specific thresholds to account for variations in participant numbers across trial phases.

From our trial knowledge base, we developed several sources of labels, including large language model (LLM) interpretations of PubMed abstracts, trial linkage across phases, sentiment analysis of news headlines, p-values, amendments, and other clinical trial metrics which were integrated into our clinical trial annotation pipeline as shown in Figure~\ref{fig:ctod1}C. 
Further details on these labeling functions are given in the supplementary. Our pipeline is validated against several baselines, including a \textit{Majority Vote}--which takes the majority vote over LFs, \textit{Data Programming}--an unsupervised matrix completion approach that weighs LFs based on mutual agreement, and a \textit{supervised Random Forest} model trained on LFs and a set of public labels (TOP).

\section*{Results}






\subsection*{CTO Automated Labeling Achieves High Agreement with Previous Expert Labels}
\begin{table}[t!]
\centering
\caption{Evaluation of automatically aggregated knowledge base data sources. Agreement of CTO with TOP (on TOP test split). $CTO_{MV}$ denotes Majority Vote, $CTO_{DP}$ denotes Data Programming \cite{ratner2016data}, and $CTO_{RF}$ denotes Random Forest. Note that RF \textit{actively} trained on the training human labels, whereas MV and DP are not. \textbf{Bolded} denotes best performance.}
\label{tab:dp}
\resizebox{.5\linewidth}{!}{
\begin{tabular}{llll} \toprule
Phase & Aggregation Method & F1 & $\kappa$ \\ \midrule
\multirow{3}{*}{I} & $CTO_{MV}$ & 0.726 & 0.490 \\
 & $CTO_{DP}$ & 0.870 & 0.700 \\
 & $CTO_{RF}$ & \textbf{0.913} & \textbf{0.790} \\ \midrule
\multirow{3}{*}{II} & $CTO_{MV}$ & 0.689 & 0.430 \\
 & $CTO_{DP}$ & 0.856 & 0.623 \\
 & $CTO_{RF}$ & \textbf{0.878} & \textbf{0.693} \\ \midrule
\multirow{3}{*}{III} & $CTO_{MV}$ & 0.904 & 0.606 \\
 & $CTO_{DP}$ & 0.921 & 0.582 \\
 & $CTO_{RF}$ & \textbf{0.941} & \textbf{0.710} \\ \midrule
\multirow{3}{*}{All} & $CTO_{MV}$ & 0.793 & 0.529 \\
 & $CTO_{DP}$ & 0.884 & 0.646 \\
 & $CTO_{RF}$ & \textbf{0.909} & \textbf{0.729} \\ \bottomrule
\end{tabular}

}
\end{table}

We evaluate our knowledge base and labeling framework by comparing the agreement between our automatically aggregated labels against the expert-annotated TOP test set. 
Table~\ref{tab:dp} summarizes the performance of various label aggregation methods across different trial phases. The label aggregation methods evaluated include Majority Vote (CTO\textsubscript{MV}), Data Programming (CTO\textsubscript{DP}), and Random Forest-based aggregation (CTO\textsubscript{RF}). Across all trial phases, the Random Forest-based approach consistently outperforms the other methods, achieving the highest scores in both F1 and Cohen’s kappa metrics. CTO\textsubscript{RF} achieves an F1 score of $0.909$ and a Cohen's kappa ($\kappa$) of $0.729$ when evaluated across all phases, indicating substantial agreement \cite{mchugh2012interrater}. Also, the phase-specific analysis further underscores the robustness of CTO\textsubscript{RF}. It achieves F1 scores of $0.913$, $0.878$, and $0.941$ in phases I, II, and III, respectively. The F1 score of 0.909 is much higher than previous trial outcome prediction models \cite{fu2023automated, wang2023spot}.

\subsection*{Creating a Manual Curated Trial Outcome Benchmark from CTO}
Automated labeling is not a substitute for human expertise, so we manually annotated over 2,500 challenging cases using our knowledge base to create a high-confidence gold standard set for evaluation. The process for developing this benchmark is illustrated in Figure~\ref{fig:ctod1}D.
We applied the same rule-based termination criteria and p-value threshold of less than 0.05 as used in the automated labeling step to filter out the easier cases. 
For the remaining unlabeled trials, we engaged three clinical trial experts to review relevant publications, news articles, and other reported outcomes to accurately determine the trial outcomes. In this benchmark evaluation, we compared the labels generated by the CTO automated labeling framework against the human-generated labels. 
The CTO framework demonstrated strong overall performance, achieving an F1 score of 97.1. Notably, even in challenging cases (after applying p-value filtering and status checks), where accurate labeling is more complex, the CTO framework maintained an F1 score of 72.7.


\begin{figure}[t!]
    \centering
    \includegraphics[width=1\linewidth]{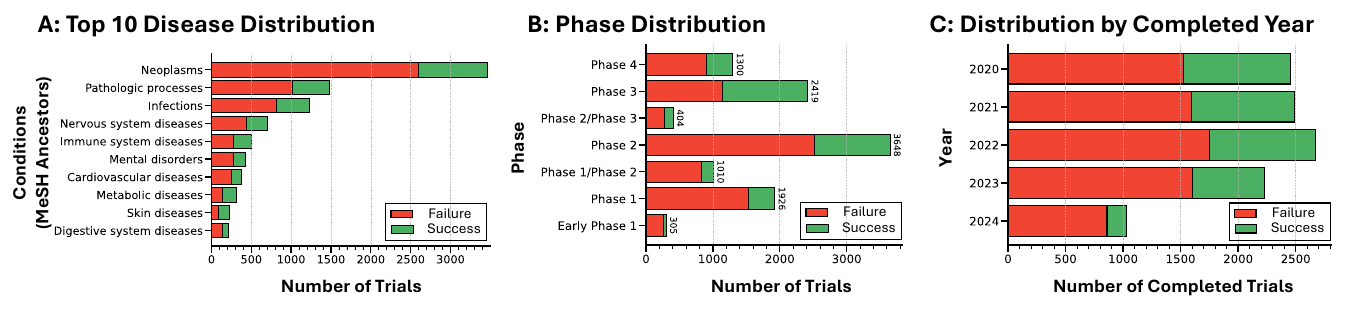}
    \caption{Overview of \textbf{our} manually curated trial outcome benchmark. (A) Bar plot illustrating the top 10 conditions in the benchmark, categorized by MeSH ancestor terms, with associated label counts. (B) Distribution of trial phases and their respective outcome labels. (C) Annual distribution of completed trials in the set from 2020-2024 and their label counts.}
    \label{fig:human_labeled}
\end{figure}

\subsection*{Consistent High Prediction Performance Over Time}
\begin{figure}[t!]
    \centering
    \includegraphics[width=1\linewidth]{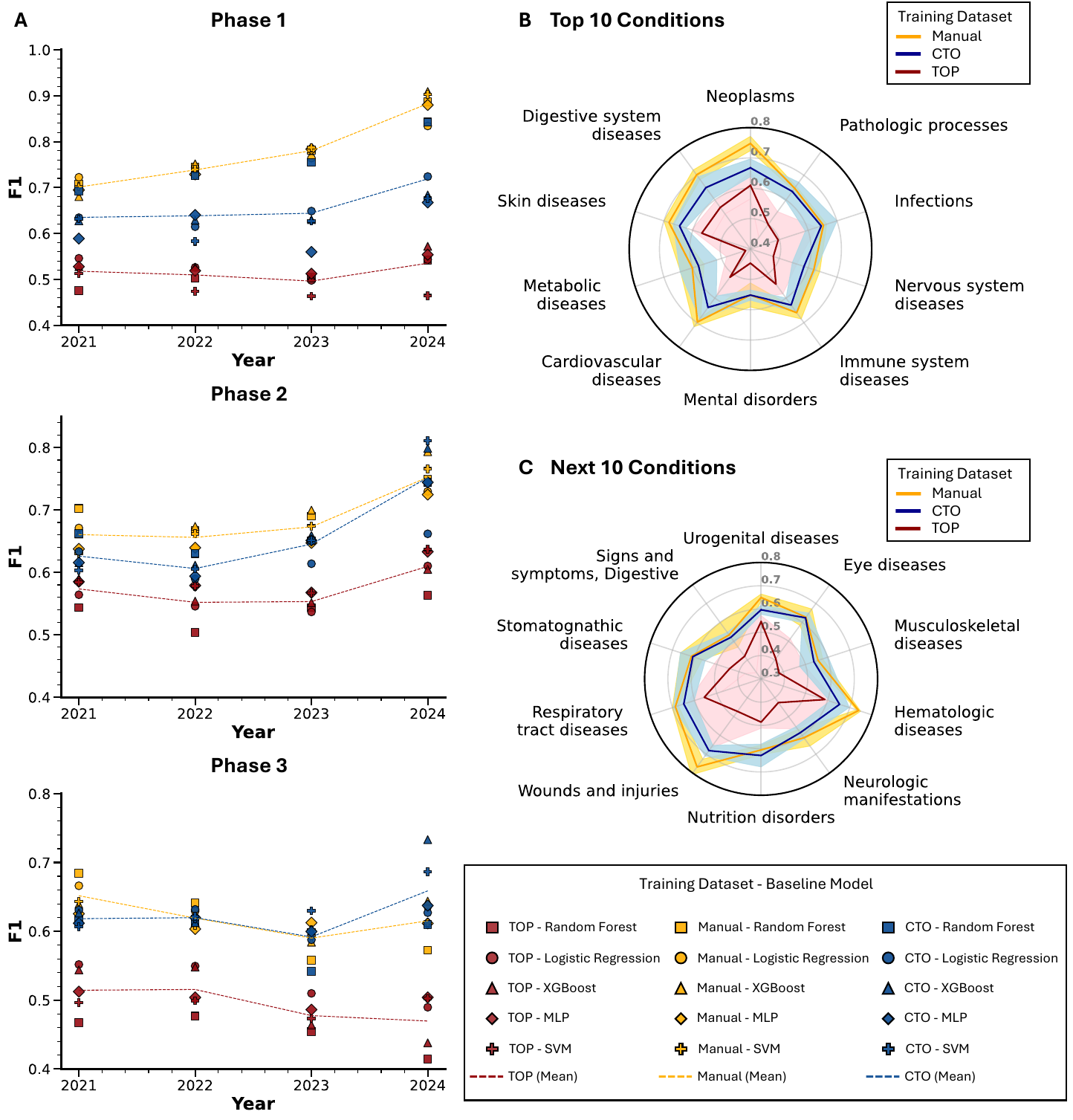}
\caption{(A) Year-by-year comparison of trial outcome prediction performance for baseline models trained on the TOP dataset, the constantly updated CTO dataset, and our manually curated set, evaluated on trials completed from 2021 to 2024. The TOP dataset is fully utilized for training, whereas for the CTO and manually curated sets, only trials completed before each test year are used for training. (B) Top 10 Condition-wise performance of baselines on manual labels, with models trained on data $<$ 2022 and tested on $\geq$ 2022 using TOP, CTO, and manually curated set. (C) Similar to B, Condition-wise performance on the next 10 (Top 11-20 conditions)   }
    \label{fig:ctovstop}
\end{figure}

Given that we have manual labels for recent trials, we conducted experiments comparing the impact of our constantly updated CTO dataset against the previously publicly available fixed benchmark TOP dataset.
We trained standard baseline trial outcome prediction models on the entire TOP dataset (train + valid + test). We assessed their year-by-year performance on trials completed between 2021 and 2024 in our manually curated dataset. For the CTO dataset, the baselines were trained using only trials completed before each test year, with training labels generated through our automated labeling framework. Similarly, we trained the baselines on the manually curated dataset using human-labeled outcomes for trials completed before each test year. The phase-wise performance comparison of these baseline models across the different training datasets is presented in Figure~\ref{fig:ctovstop}A.

Baseline models trained using labels from the CTO dataset consistently outperform their counterparts trained on the TOP dataset across all years and trial phases. The CTO dataset significantly enhances model performance compared to the TOP dataset, underscoring the importance of constantly updated labels. This improvement is mainly attributable to distribution shifts caused by evolving trial practices, therapeutic areas, and advancements in trial designs over time.


CTO provides a scalable solution for labeling recent trial data, overcoming the infeasibility and time demands of obtaining new human labels. By achieving high agreement with human labels, the CTO enables models to obtain close-to-human-level performance by leveraging large datasets to adapt to the evolving landscape of clinical trials.

\subsection*{Performance of Condition-Specific Outcome Predictions}
Figure~\ref{fig:ctovstop}B and C present the performance of baseline models trained using our human-labeled data across the top 20 conditions in the dataset (3B shows conditions 1-10, and 3C shows conditions 11-20). Here, we train the baselines on trials completed before 2022 over TOP, CTO, and manually curated. The baselines are tested on the manually curated trials completed after 2022. The performance of the baseline models trained on CTO closely follows the models trained on our human labels. The performance demonstrates that the labels produced by CTO align closely with manual annotations despite being generated independently through an automated process. Moreover, these CTO-based models consistently outperform those trained on the TOP dataset, demonstrating the value of using a more current dataset. Furthermore, CTO facilitates the development of condition-specific clinical trial outcome prediction models by rapidly generating automated labels for condition-specific datasets, significantly reducing the time and effort involved in manual labeling, with CTO-trained baselines matching or even exceeding baselines trained on manually-curated in some cases.



\section*{Discussion}


\begin{figure}[t!]
    \centering
    \includegraphics[width=1\linewidth]{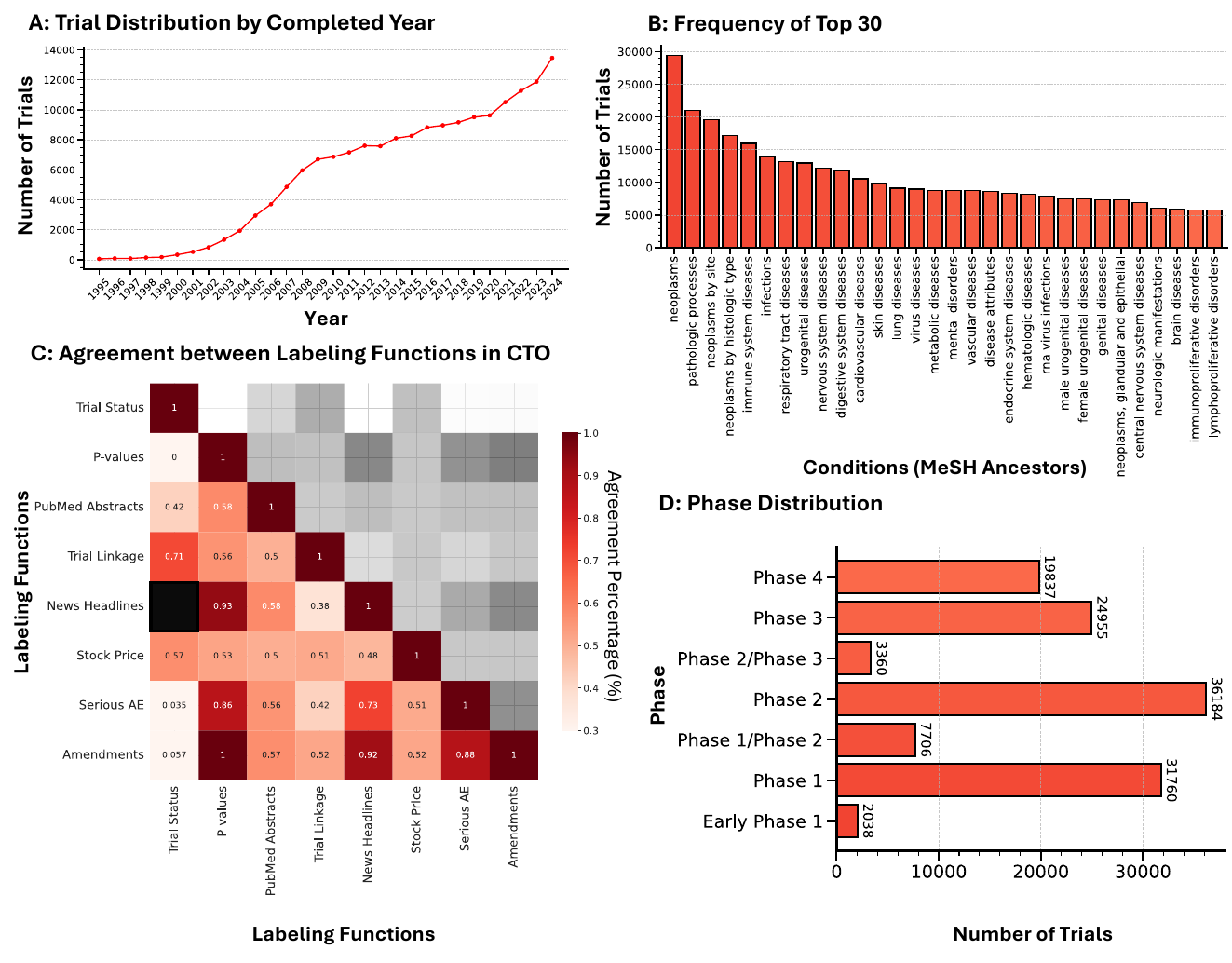}
    \caption{Overview of the CTO dataset and labeling functions. (A)Trial distribution by completed year in CTO, which shows a steady increase over the years. (B) Frequency of top 30 conditions in the CTO dataset, categorized by MeSH ancestor terms. (C) Heatmap shows the agreement between the important labeling functions in CTO. (D) Distribution of trial phases and their counts in CTO.}
    \label{fig:cto_statistics}
\end{figure}
The CTO dataset provides a comprehensive view of clinical trial data, capturing key information across multiple dimensions for around ~125K drug and biologics trials. Figure~\ref{fig:cto_statistics}A, B, and D provide an overview of the dataset's key statistics, highlighting its scale and diversity. Figure~\ref{fig:cto_statistics}A illustrates the annual distribution of completed clinical trials, and it shows a steady increase over the years, reflecting the growing scale of global clinical research. In Figure~\ref{fig:cto_statistics}B, we show the frequency distribution of the top 30 medical conditions within the dataset, where Neoplasms are the most represented condition.
The phase distribution of trials in the CTO dataset is shown in Figure~\ref{fig:cto_statistics}D. 
Note that we ensured that our set of manual labels accurately reflects the overall clinical trial distribution as shown in Figure~\ref{fig:human_labeled}A, B, and C.

\subsection*{How Well the Labeling Functions Agree?}

To assess the agreement between the labeling functions, we calculated the percentage of agreement for trials that fall within the intersections of their coverage, as shown in Figure~\ref{fig:cto_statistics}C. The heatmap highlights the agreement scores only for the important labeling functions in the dataset. Each cell in the heatmap represents the agreement score between a pair of labeling functions. It is important to note that each labeling function does not cover all trials with weak labels. Therefore, we only considered the trials common to both labeling functions while calculating the agreement score. Because of this, there are no common trials with weak labels between the news headlines and trial status labeling functions (represented as a black cell). This is because the trial status labeling function assigns a value of -1 (indicating an inability to decide the trial outcome) to trials with a `completed' status, while news extraction is limited to `completed' trials to optimize resource usage associated with SerpAPI.

From the heatmap in Figure~\ref{fig:human_labeled}F, several key patterns in labeling function agreement can be observed.  Notably, there is a high level of agreement between news headlines and labels derived from p-values, amendments, and serious adverse events. Additionally, we observe a strong agreement between trial linkage and trial status, suggesting that trial linkage effectively identifies failure trials, as most trial status-based labels pertain to trials categorized as terminated,' withdrawn,' or `suspended.' 

Another noteworthy observation is the low agreement between p-values and trial status. This discrepancy arises because most trial status-based failure labels correspond to trials marked as `terminated,' `withdrawn,' or `suspended,' often due to sponsor or business decisions rather than any trial outcome. Interestingly, some of these trials report statistically significant outcomes (defined as p-values($<0.05$) for at least one of their primary endpoints but were stopped for reasons unrelated to efficacy. For instance, our analysis identified approximately 1,200 trials marked as ``terminated'' that reported at least one statistically significant primary outcome. Specific examples include NCT04364763, which was terminated due to enrollment feasibility issues, and NCT01374451, which was stopped for failing to meet the primary endpoint of progression-free survival. These cases highlight the complex interplay between trial outcomes and external factors.

\section*{Conclusion}



We present the first large-scale, publicly available knowledge base for clinical trial outcomes, providing automatically generated outcome labels alongside a manually curated dataset annotated by human experts. By publicly releasing our trial annotation pipeline and the corresponding knowledge base, we aim to democratize trial outcome prediction, fostering more transparent and data-driven clinical trial design. Our dataset, including recent trials, weak labels, and the full knowledge base, is available at \url{https://chufangao.github.io/CTOD/}, with regular updates to support ongoing research and innovation.



\paragraph{Acknowledgement}
This work was supported by NSF award SCH-2205289, SCH-2014438, and IIS-2034479.
\bibliography{custom}
\bibliographystyle{naturemag}

\appendix

\clearpage
\section*{Ethics and Broader Impacts}
Using a weakly supervised dataset for clinical trial outcome prediction with large language models (LLMs) can potentially decrease the reliability of the model's predictions if not correctly instantiated with proper labeling functions. Weak supervision may result in incomplete or imprecise labeling, leading to the model learning incorrect associations and missing crucial factors, which can introduce or exacerbate biases. This lack of precise guidance can also cause the model to overfit to noise or incorrect patterns in the training data, reducing its ability to generalize effectively to new, unseen data. In diverse clinical scenarios, one must take care to independently verify model predictions to prevent potentially jeopardizing patient safety with inaccurate predictions. To mitigate these issues, it is essential to improve the quality of the training data through better labeling techniques, supplementary high-quality data, or advanced methods like semi-supervised or active learning.

Furthermore, we use publicly available data, so the risk of identification is minimized. However, we recognize that LLMs on public data inherently still pose some privacy risks.

\paragraph*{Reproducibility}
We utilized an AMD EPYC 7513 32-core Processor with 100 GB of RAM to run our experiments. Running Data Programming with less or more powerful systems may impact speed. For us, obtaining the labels took around 25 hours in experiments and prototyping. Additionally, we made use of ChatGPT for writing as well as obtaining trial outcome prediction on the PubMed abstracts. Total cost was around \$200 US dollars.

\section*{Related Work}

\begin{table}[ht]
\centering
\caption{This table describes an overview of selected recent trial outcome prediction work. All previous work relies on industry sponsors to obtain labels, and most are not easily accessible. We are the first to aggregate publicly available data sources on a large number of trials.}
\label{tab:prevwork}
\Huge
\resizebox{\linewidth}{!}{
\begin{tabular}{llllcc} 
\toprule[1pt]
\textbf{Dataset} & \textbf{Data Sponsor} & \textbf{Subset} & \textbf{\# Trials} & \textbf{Label Creation} & \begin{tabular}[c]{@{}l@{}}\textbf{Publicly} \\ \textbf{Available}\end{tabular} \\ \midrule
Lo et al. \cite{lo2019machine} & TrialTrove, Pharmaprojects & Phase II, III & 19,136 & Human Expert & \xmark \\
Feijoo et al. \cite{feijoo2020key} & Biomedtracker & Industry Phase II, III & 6,417 & Manual linking & \xmark \\
Aliper et al. \cite{aliper2023prediction} & Insilico Medicine & Phase II & 55,653 & Human Expert & \xmark \\
Willigers et al. \cite{willigers2023algorithmic} & AstraZeneca & AstraZeneca Phase III & 57 & Human Expert & \xmark \\
TOP \cite{fu2022hint} & IQVIA & Small Molecule Drugs & 17,538 & Human Expert & \cmark \\ \midrule
\method\ (ours) & Publicly Collected & All & 125,840 & \begin{tabular}[c]{@{}l@{}}Automatic Aggregation \\ (Tuned by Human Labeling)\end{tabular} & \cmark \\ \bottomrule[0.8pt]
\end{tabular}
}
\end{table}

Predicting clinical trial outcomes is often led by industries with the capacity for extensive data curation. Informa's TrialTrove \cite{stergiopoulos2019evaluating} is widely used, containing around 20,000 trials \cite{lo2019machine}. AstraZeneca has developed structured feedback forms to improve Phase 3 trial success annotations \cite{willigers2023algorithmic}. Feijoo et al. \cite{feijoo2020key} demonstrated that using Random Forest on Biomedtracker, a proprietary dataset aggregating company reports, results in strong outcome prediction performance.

Previous studies have tackled clinical trial outcome prediction using various methods. Early work employed statistical analysis \cite{malik2014predicting} and ML models on limited private data sets ($<$500 samples) \cite{dimasi2015tool}. In drug toxicity prediction, Gayvert et al. \cite{gayvert2016data} used Random Forest to predict outcomes based on chemical properties. In contrast, Artemov et al. \cite{artemov2016integrated} used Multi-layer Perceptrons (MLP) for Phase I/II trials. Lo et al. \cite{lo2019machine} applied k-NN imputation and Random Forest on features from Pharmaprojects and TrialTrove. Additionally, Phase 2 to Phase 3 prediction was explored by Qi et al. \cite{qi2019predicting} through clinical trials and by Aliper et al. \cite{aliper2023prediction} using experts, GPT-3.5, and a biomedical knowledge graph.

Previous works have addressed clinical trial outcome prediction in various ways, with early work focusing on statistical analysis \cite{malik2014predicting} and ML models on small amounts of private outcome labels \cite{dimasi2015tool} (<500 samples).
A related field is drug toxicity prediction, where Gayvert et al. \cite{gayvert2016data} used Random Forest to predict drug toxicity outcomes from chemical properties and target-based features including molecular weight and polar surface area. Similarly, Artemov et al. \cite{artemov2016integrated} used Multilayer Perceptions to predict the toxicities and outcomes of Phase I/II drugs clinical trials. 

Lo et al. \cite{lo2019machine} used K-nearest-neighbor (KNN) imputation and Random Forest (RF) to predict outcomes on features from two commercial databases, Pharmaprojects and Trialtrove. 
Additionally, Phase 2 to Phase 3 prediction has been explored \cite{qi2019predicting, aliper2023prediction}, However, Qi et al. \cite{qi2019predicting} conducted their clinical trial to obtain the data, and Aliper et al. \cite{aliper2023prediction} used experts, GPT3.5, as well as a proprietary biomedical knowledge graph to annotate their transition success labels for phase 2 trials. 

Fu et al. \cite{fu2023automated} released the first \textit{and only} publicly available clinical trial outcome dataset based on manual curation. Train and test data splits were selected as all trials completed before 2014 and afterward respectively. After preprocessing and cleaning, the final number of trials in train, validation, and test datasets was reduced to 17,538 $\rightarrow$ 12,465 trials, which was only around 4\% of all available trials at the time of publication \footnote{Note that the full set of small-molecule drug interventional trials, which TOP primarily focuses on, consists of around 100k trials}. Table~\ref{tab:prevwork} summarizes recent trial outcome work.




\section*{Benchmarking Baselines on Manually Curated Trial Outcome Benchmark}

\begin{table}[t!]
\centering
\caption{Table of performance of baseline model on only the human-label subset. Training data were trials that had completion dates $<2022$ and test data were trials that had completion dates $\geq 2022$.}
\label{tab:humanlabel}
\resizebox{.5\linewidth}{!}{
\begin{tabular}{cclll} \toprule
Phase              & Model      & F1                 & PRAUC              & ROCAUC             \\ \midrule
\multirow{7}{*}{1} & RF         & 0.790$_{\pm0.011}$ & 0.743$_{\pm0.015}$ & 0.544$_{\pm0.011}$ \\
                   & LR         & 0.750$_{\pm0.010}$ & 0.766$_{\pm0.013}$ & 0.585$_{\pm0.015}$ \\
                   & XGBoost    & 0.786$_{\pm0.011}$ & 0.739$_{\pm0.014}$ & 0.546$_{\pm0.012}$ \\
                   & MLP        & 0.771$_{\pm0.010}$ & 0.748$_{\pm0.013}$ & 0.562$_{\pm0.013}$ \\
                   & SVM        & 0.781$_{\pm0.010}$ & 0.741$_{\pm0.012}$ & 0.546$_{\pm0.012}$ \\
                   & BioBERT    & 0.745$_{\pm0.010}$ & 0.763$_{\pm0.014}$ & 0.556$_{\pm0.015}$ \\
                   & PubMedBERT & 0.761$_{\pm0.011}$ & 0.758$_{\pm0.015}$ & 0.546$_{\pm0.014}$ \\ \midrule
\multirow{7}{*}{2} & RF         & 0.675$_{\pm0.009}$ & 0.652$_{\pm0.011}$ & 0.581$_{\pm0.012}$ \\
                   & LR         & 0.646$_{\pm0.008}$ & 0.655$_{\pm0.01}$  & 0.579$_{\pm0.010}$  \\
                   & XGBoost    & 0.696$_{\pm0.009}$ & 0.665$_{\pm0.012}$ & 0.585$_{\pm0.011}$ \\
                   & MLP        & 0.653$_{\pm0.009}$ & 0.658$_{\pm0.012}$ & 0.567$_{\pm0.012}$ \\
                   & SVM        & 0.683$_{\pm0.009}$ & 0.650$_{\pm0.010}$ & 0.571$_{\pm0.011}$ \\
                   & BioBERT    & 0.659$_{\pm0.009}$ & 0.651$_{\pm0.011}$ & 0.567$_{\pm0.011}$ \\
                   & PubMedBERT & 0.670$_{\pm0.010}$ & 0.636$_{\pm0.011}$ & 0.550$_{\pm0.011}$  \\ \midrule
\multirow{7}{*}{3} & RF         & 0.581$_{\pm0.013}$ & 0.607$_{\pm0.011}$ & 0.606$_{\pm0.011}$ \\
                   & LR         & 0.601$_{\pm0.015}$ & 0.603$_{\pm0.012}$ & 0.615$_{\pm0.014}$ \\
                   & XGBoost    & 0.584$_{\pm0.012}$ & 0.594$_{\pm0.011}$ & 0.597$_{\pm0.011}$ \\
                   & MLP        & 0.599$_{\pm0.014}$ & 0.581$_{\pm0.012}$ & 0.607$_{\pm0.014}$ \\
                   & SVM        & 0.589$_{\pm0.013}$ & 0.587$_{\pm0.011}$ & 0.600$_{\pm0.012}$   \\
                   & BioBERT    & 0.588$_{\pm0.012}$ & 0.560$_{\pm0.012}$ & 0.588$_{\pm0.012}$ \\
                   & PubMedBERT & 0.601$_{\pm0.013}$ & 0.567$_{\pm0.012}$ & 0.599$_{\pm0.013}$ \\ \bottomrule
\end{tabular}
}
\end{table}


Table~\ref{tab:humanlabel} summarizes the performance of various baseline models on our CTO gold labels, where training was conducted using our manually labeled gold set trials completed before 2022, and testing was done on trials completed post-2022. Performance metrics include F1 score, Precision-Recall Area Under the Curve (PRAUC), and Receiver Operating Characteristic Area Under the Curve (ROCAUC). Each performance measure is weighted by label prevalence to ensure a fair comparison of both positive and negative classes. 

RF denotes random forest, LR denotes logistic regression, MLP denotes multilayer perception with 1 hidden layer of dimension 100, and SVM is the Support Vector Machine linear classifier. The input features are the term-frequency times inverse document-frequency (TFIDF) transformation of the concatenated trial phase, disease, interventions, design, and inclusion/exclusion criteria. 
For BioBERT and PubMedBERT, we use both of these large language models (LLMs) to encode the input text (instead of TFIDF) to obtain embedding features on which a 2-layer MLP with hidden dimension 100 is trained. 

Advanced trial outcome models, such as HINT \cite{fu2022hint} and SPOT \cite{wang2023spot}, present notable limitations that hinder their adaptability to newer clinical trials. Both approaches rely on pretrained models to predict ADMET properties—absorption, distribution, metabolism, excretion, and toxicity. However, in our evaluations, these pretrained models demonstrated suboptimal adaptability when applied to novel drugs and biologics, likely due to domain shifts or insufficient representation of emerging compounds. While optimizing these sophisticated techniques is highly valuable, our primary focus in this study is the development and curation of the dataset, laying the groundwork for better trial modeling for future work.

In Phase 1, models demonstrated strong overall performance, with Random Forest (RF) achieving the highest F1 score (0.79), closely followed by XGBoost (0.786) and Support Vector Machines (SVM, 0.781). Logistic Regression (LR) excelled in terms of precision-recall trade-offs, reflected in its highest PRAUC (0.766) and ROCAUC (0.585) scores, suggesting superior discriminative ability in this phase. 
Interestingly, large language models--BioBERT and PubMedBERT-- underperformed slightly in F1 scores. This could be due to the possible overfitting on the training set, as these models are much larger than the classical ones, and we see in the original paper that distribution shift is an issue over each consecutive year.

In Phase 2, performance declined across all models, suggesting increased complexity or noise in the dataset. XGBoost emerged as the most robust model, achieving the highest F1 score (0.696), PRAUC (0.665), and ROCAUC (0.585). RF (F1: 0.675) and SVM (F1: 0.683) also showed resilience, maintaining consistent performance. However, deep learning models exhibited a pronounced decline, with PubMedBERT and BioBERT achieving lower F1 scores (0.67 and 0.659, respectively) and showing limited adaptation to the challenges of this phase. These findings emphasize the robustness of XGBoost and the potential limitations of pre-trained transformer-based models when applied to datasets with phase-specific complexities.

Phase 3 presented the greatest challenges, with performance metrics reaching their lowest levels. Logistic Regression demonstrated the most robust performance, achieving the highest F1 score (0.601) and ROCAUC (0.615), indicative of its adaptability to the phase's complexities. PubMedBERT performed comparably in terms of F1 (0.601) but lagged in precision-recall balance, as evidenced by its lower PRAUC (0.567). Classical models such as RF and XGBoost performed moderately but did not exhibit the same dominance seen in earlier phases. Deep learning models struggled consistently across all metrics in Phase 3, suggesting that their reliance on pre-trained features may not generalize effectively to this phase's data distribution.

Across all phases, a clear trend of declining performance from Phase 1 to Phase 3 was observed, potentially due to increasing data complexity, reduced signal strength, or distributional shifts in the features. Classical models, particularly XGBoost and RF, were robust in Phases 1 and 2, while Logistic Regression excelled in Phase 3, highlighting the difference in data distribution between each phase. Deep learning models underperformed overall, indicating the need for additional fine-tuning or integration with domain-specific adaptations to achieve comparable results. These findings suggest that ensemble methods combining the strengths of RF, XGBoost, and LR could improve overall performance. 
Future work should investigate feature importance and conduct error analysis to identify misclassification patterns, guiding model refinement. 
This table also underscores the need for a phase-aware modeling approach to optimize clinical trial outcome prediction.

\begin{figure}[!t]
    \centering
    \includegraphics[width=0.85\linewidth,trim={0.5cm 2.5cm 0.5cm 0cm},clip]{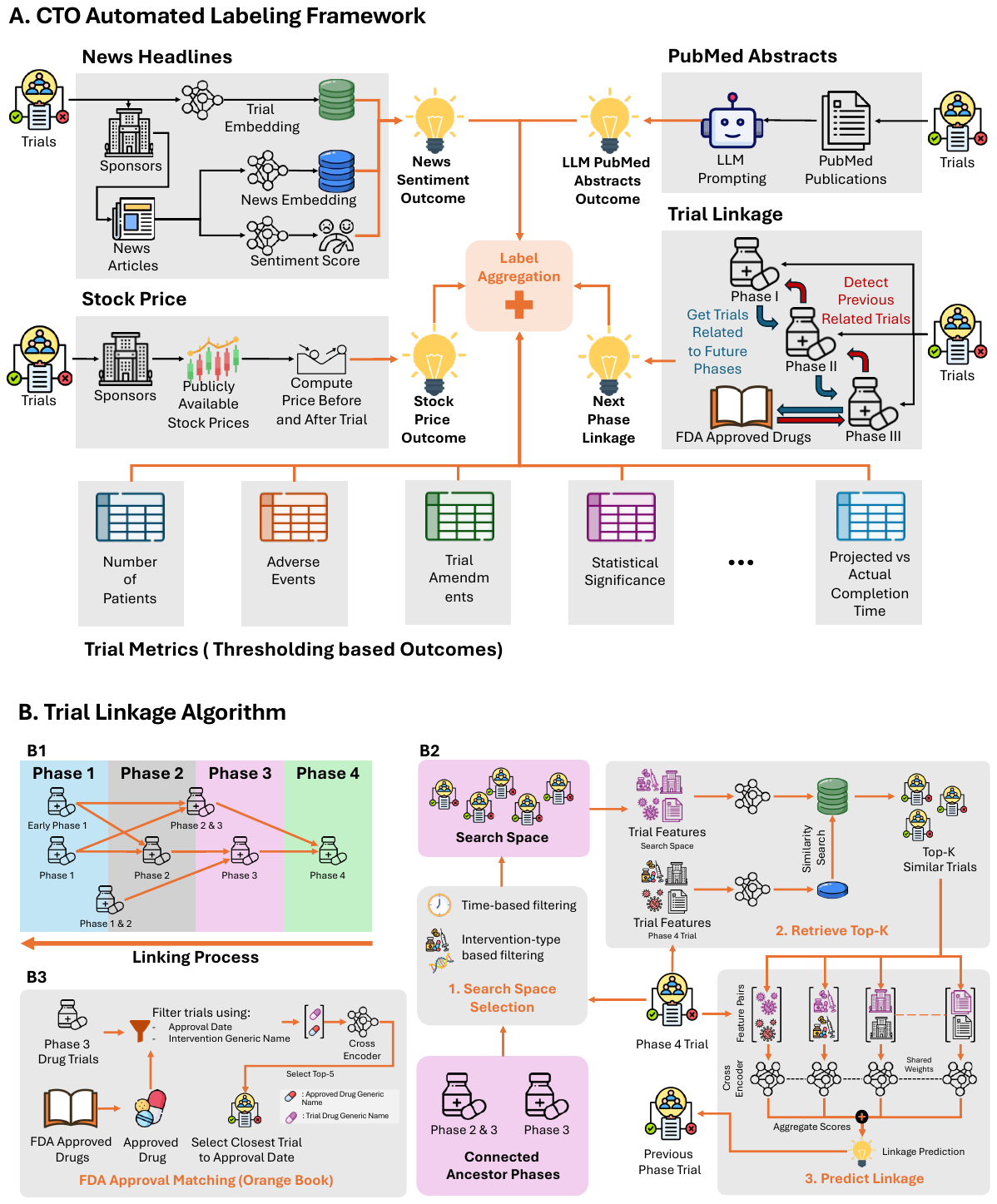}
    \caption{(A) provides the overview of the \method automated labeling framework. \method integrates various data sources to generate labels for predicting clinical trial outcomes. Data sources include (1) News articles, from which trial embeddings and sentiment scores are derived; (2) Publicly available stock prices, used to compute price changes before and after trials; (3) PubMed abstracts, where LLMs are prompted to predict the trial outcome; and (4) trial phase linkage. Additional sources of outcomes include trial metrics such as the number of patients, adverse events, etc. (B) Overview of trial linkage algorithm. (B1) Illustrates the phase connection map, covering all the phase categories present in the dataset. 
    (B2) Linking trials across phases based on completion dates and intervention types, disease, and other trial data. We 1) constraint the search space, then 2) retrieve top-K most similar past trials using embeddings similarity. Finally, 3) a cross-encoder re-ranking strategy predicts linkages by scoring likely pairs.
    (B3) Matching FDA-approved drugs from FDA orange book to Phase 3 and Phase 2 $\&$ 3 drug trials.}
    \label{fig:trial_linkage} 
    
\end{figure}

\section*{\method\ Overview}
\label{sec:overview}
Our main methodology is outlined in Figure~\ref{fig:trial_linkage}A. We overview our primary sources of outcome predictors--LLM prediction on Pubmed Abstracts, Trial Linkage, News Headlines, Stock Price, and finally, Trial Metrics computed from clinicaltrials.gov. These metrics are computed independently and are aggregated via weakly supervised label aggregation.

\subsection*{LLM Predictions on PubMed Abstracts}
\label{sec:gpt}
As many trials had multiple abstracts, we selected the top 2 abstracts based on their title similarity to the trial's official title to provide the most relevant information. Given these abstracts, we prompted the `GPT-3.5-turbo' model to summarize important trial-related statistical tests and predict the outcome. 



\subsection*{Trial Linkage}
\label{sec:linkage}
The journey of a drug from discovery to FDA approval involves several stages, beginning with Phase 1 trials to assess safety and dosage. Subsequent Phase 2 and 3 trials evaluate efficacy and compare the new drug to existing therapies. Upon completing Phase 3, a drug may be submitted for FDA approval. A key limitation of the CTTI dataset is the lack of connectivity between trial phases, which could significantly enhance the ability to analyze trial progression and outcomes based on advancement to subsequent phases. Moreover, linking trials across phases is not straightforward due to challenges, including unstructured data, inconsistent reporting standards, missing information, data noise, and discrepancies in intervention details across phases. Despite these challenges, linking trials can be invaluable, particularly as a source of weak labels in clinical trial outcome prediction tasks. This section presents our novel trial-linking algorithm as illustrated in Figure~\ref{fig:trial_linkage}B, which, to our knowledge, is the first attempt to systematically connect different phases of clinical trials.
The trial linkage extraction process consists of two primary steps: 1) Linking trials across different phases and 2) Matching phase 3 trials with FDA approvals (Figure~\ref{fig:trial_linkage}B3).

The progression of clinical trials through phases is not all strictly sequential from Phase 1 to Phase 3, as some studies may combine multiple phases. Many trials were categorized as `Not Applicable' or were missing phase information entirely; these were excluded from our analysis. We created a phase connection map, illustrated in Figure~\ref{fig:trial_linkage}B1, that covers all phase categories present in the dataset. Our linking algorithm begins with the later phases and traces back to earlier phases.  This approach is based on the assumption that a trial in a subsequent phase implies the success and existence of a corresponding trial in a preceding phase. Phase 3 trials are considered a success if a Phase 4 trial is found or if it is linked to an FDA new drug application.

The trial linkage process consists of three main steps: 1) Selection of the search space, 2) Retrieval of the top K most similar past trials, and 3) Prediction of linkage.  In Figure~\ref{fig:trial_linkage}B2, we illustrate an example of linking a Phase 4 trial to its preceding phases. For a given trial $x$, the objective is to identify its predecessor among trials in its directly linked earlier phases. For instance, the directly linked earlier phases of Phase 4 are Phase 3 and Phase 2$\&$3. 

%

\textbf{1. Search space selection:} We filter the trials based on their completion dates relative to the start date, ensuring that any linkage candidate must have concluded before the start of $x$. Furthermore, we also consider intervention types such as `Drug', `Biological', `Device', etc. 

\textbf{2. Retrieve top-K:} From the filtered search space $\mathbf{Z}$, we retrieve top-32 most similar past trials to $x$. We extract key features and encode them into dense embedding using PubMedBERT \cite{gu2021domain} to represent both $x$ and trials in the search space ($z^i \in Z$) as follows: $x = \{x_{\text{I}},x_{\text{C}},x_{\text{T}},x_{\text{S}},x_{\text{E}}\}$,  $z^i =  \{z^i_{\text{I}},z^i_{\text{C}},z^i_{\text{T}},z^i_{\text{S}},z^i_{\text{E}}\}$.
Where the subscript $I$ denotes intervention or drug, $C$ denotes condition or targeted disease, $T$ denotes official trial title, $S$ denotes trial summary, and $E$ denotes eligibility criteria. We calculate similarity as: $\text{similarity}(x^i,z^i) = \sum_{j \in \mathbf{F}}\frac{x^i \cdot z_j^i}{\|x^i\| \|z_j^i\|}
$,
where $\mathbf{F} =$ \{I,C,T,S,E\}.
    
We excluded the lead sponsor as a feature since the sponsor often changes depending on funding and performs worse empirically (See Supplementary: Ablation on Features for Trial Linkage).

\textbf{3. Predict linkage} Given the large search space, we employ a re-ranking strategy using a cross-encoder pre-trained on MS-MARCO \cite{bajaj2016ms}. We provide feature pairs as input to the cross-encoder as follows: $\text{Cross-encoder score}(x^i,z^i) = \sum_{j \in \mathbf{F}}g_{\theta}(x^i,z^i_j)$.
Based on the cross-encoder scores, we predict the linkage by considering the trials with the highest positive cross-encoder scores as the most probable previous phase trials of $x$.

We apply this process for all trials in Phase 4, Phase 3, Phase 2 \& 3, and Phase 2 to extract the trial linkages. To extract the outcome labels, we start with trials in the earlier phases and label them based on the existence of linked trials in the subsequent phases. However, trials in Phase 3 and Phase 4 have some exceptions to this process. For Phase 4 trials, there are no following trial phases, so we exclude them from the extracted weak labels. As for Phase 3 trials, they can be successful even without the existence of a subsequent Phase 4 trial. This highlights the importance of matching Phase 3 trials with their corresponding FDA approvals if they exist.

After establishing connections across different phases of clinical trials, we focus on matching the Phase 3 trials to drug approvals to obtain their outcome labels. We utilize the FDA Orange Book\footnote{https://fda.gov/drugs/drug-approvals-and-databases/orange-book-data-files} version as of April 2024. Specifically, we use the approval date and drug name provided in the `product.txt' file, as the other files do not contain the relevant information required for the matching process. In this process, we only consider drug-related trials in Phase 3 and Phase 2 \& 3 since the Orange Book solely comprises FDA-approved drugs. 
For a given FDA-approved drug, we first filter Phase 3 and Phase 2 \& 3 trials based on the approval date and intervention generic name, retaining trials completed between 2 years and 2 months prior to the approval date to align with the FDA approval process timeline. Similar to trial linkage prediction, we provide the drug's generic name and the trial's intervention generic name as input pairs to a cross-encoder, predicting their similarity. We select the top 5 trials based on cross-encoder scores and match the FDA approval to the trial with the completion date closest to the approval date. We then update the previously extracted outcomes for Phase 3 and Phase 2 \& 3 trials, labeling them as successful if matched to an FDA approval or having a linked Phase 4 trial.

\subsection*{News Headlines}
\label{sec:news}
News headlines were obtained via the following steps: 
We sent requests to Google News for all completed drug and biologic trials, which accounted for around 85k trials. Due to rate limitations, we utilize an external API (SerpAPI). We found that most trials do not have corresponding news--the total number of trials that had any corresponding news articles was 3688.

We utilize Twitter-roBERTa-base for Sentiment Analysis \cite{loureiro2022timelms} to obtain sentiment (`Positive', `Negative', or `Neutral' with a confidence score between 0 and 1) for every headline.

For the final labeling function prediction, we predicted 1 if the mode of all corresponding news was `Positive' or `Neutral', 0 if the mode was `Negative', and -1 (abstain) otherwise. From our perspective, if any news article was written about the trial, it should be relatively obvious if it was good or not. Often, news is written regarding trial plans or derivative results of a trial. In such cases, we make the assumption that such trials are more likely to be successful, as they are more ``exciting'' or are created from good results from a previous trial of the same drug.


\subsection*{Stock Price} \label{sec:stock}
In this section we provide more details on how the extracted stock price information is matched with the trials. By averaging the stock prices over a specified time frame, the moving average reduces the noise caused by random, short-term price movements, making it easier to identify the underlying trend \cite{investopedia_moving_average, schwab_momentum_indicator}. A 5-day simple moving average (SMA) of a stock's price is calculated by taking the average of the closing prices for the last 5 trading days. As shorter periods reveal shorter-term trends, we selected a 5-day SMA to capture the immediate short-term impact of the completion of clinical trials. A positive slope indicates an uptrend, while a negative slope indicates a downtrend in the SMA line \cite{schwab_momentum_indicator}. The absolute value of the slope represents the steepness of the trend. We calculated the slope for a 7-day window starting at a clinical trial's `completion date'. 

\subsection*{Clinical Trial Metrics} 
We utilize the CTTI dataset\footnote{\url{https://aact.ctti-clinicaltrials.org}} to obtain preprocessed tables of trial details (e.g. eligibility criteria, statistics, linked references \cite{califf2012characteristics,tasneem2012database,anderson2015compliance}.
Specifically, we utilize the following information. (1) Whether results were reported, (2) The number of sponsors, (3) The number of patients, (4) The patient dropout rate, (5) The number of sites or locations for the trial, (6) Whether the P-value $<$ 0.05, (7) The date at which the trial was last updated vs its completion date, (8) Number of deaths, (9) Number of serious adverse events, (10) Number of \textit{any} adverse events, (11) The status of the trial e.g. terminated/withdrawn/completed/etc, and finally, (12) The number of amendments made to the trial page. Please see Weakly Supervised Labeling Functions in the Appendix for a discussion on how trial outcomes are linked to these metrics. 

Each of these metrics is treated as a weakly supervised Labeling Function (LF). 
For most of these metrics, we consider the "good" outcome as having greater or less than the median of that metric. For example, in the \textit{Serious Adverse Events LF}, we predict "1" if a trial's number of serious adverse events is less than the overall median number of serious adverse events. Otherwise, we predict "0". 
While choosing a highly specific threshold could be better than choosing the median, we note that LFs in the data programming framework (as we describe in the next section) do not have to be perfect, only better than random for data programming to work well.

\section*{Label Creation Continued}

\subsection*{Data Programming}
\label{app:dp}
The full data programming framework is detailed by Ratner et al. \cite{ratner2019training}. We introduce a small aspect of the framework below.
At a high level, the aggregation of weakly supervised labeling functions (LFs) is framed as a dependency graph $G_{source}$ where each LF $\lambda_i$ is dependently conditioned on the true label $Y$. In our case, we assume conditional independence of all $\lambda_i | Y$. For this case, the dependency graphs will have observable cliques  $\bm{O} = \{\lambda_i, i \in n_{lf}\} \subset C$, where $n_{lf}$ is the number of labeling functions.

From here, the covariance matrix of an observable subset of the cliques in $G_{source}$ is analyzed, leading to a matrix completion approach for recovering estimated accuracies $\mu$ (used in the final label model to predict $P(\bm{Y}|\bm{\lambda})$).

Let $\mu = \mathbb{E}(\psi(C))$ where $\psi(C)$ is a vector of indicator random variables for all combinations of all but one of the labels emitted by each variable in clique C.

The norm of the covariance of observed LFs cliques $O$ and separator set $S$ cliques $\bm{Cov}(\psi(O) \cup \psi(S))$ can be used to recover $\mu$. 

\begin{align}
& \bm{Cov}(\psi(O) \cup \psi(S)) = \Sigma =  \begin{bmatrix}
\Sigma_{O} & \Sigma_{OS} \\
\Sigma_{OS}^T & \Sigma_{S}
\end{bmatrix}
\end{align}
Its inverse is:
\begin{align}
& K = \Sigma^{-1} =  \begin{bmatrix}
K_{O} & K_{OS} \\
K_{OS}^T & K_{S}
\end{bmatrix}
\end{align}
Applying block matrix inversion, we get:
$$K_O = \Sigma^{-1}_O + c \Sigma^{-1}_O \Sigma_{OS} \Sigma^T_{OS} \Sigma^{-1}_{O}$$
$$c = (\Sigma_{S} - \Sigma^T_{OS} \Sigma^{-1}_{O} \Sigma_{OS})$$
Let $z = \sqrt{z} \Sigma^{-1}_{O} \Sigma_{OS}$, then
$$K_O = \Sigma^{-1}_O + zz^T$$
Solving for $z$ can directly recover estimated accuracies from $\mu$ via Algorithm 1 in Ratner et al. \cite{ratner2019training}.



\subsection*{Weakly Supervised Labeling Functions}
\label{app:metrics}
In this section, we report all of our LFs and their implementations.

\begin{itemize}
\item \texttt{results\_reported}: 1 if results were reported for a trial. Otherwise, it predicts 0. 

\item \texttt{num\_sponsors}: The number of sponsors for a trial. Can be thresholded.  It is important to study the impact of single sponsors, multiple sponsors, collaborative partnerships, and public funding. The number and type of sponsors will have a significant impact on the clinical trial process and the overall path of bringing the new drug to market. Sponsors affect all parts of the trial, from funding and resources, regulatory guidance, global reach, operational support, supply chain management, market access and distribution, and risk management.  

\item \texttt{num\_patients}: The number of patients for a trial. Can be thresholded. As the clinical trials are conducted with strong statistical power to detect the true effect of the drug and to minimize the risk of committing Type II error (failing to detect the treatment effect that is present or false negatives), number of participants in each respective trial is key aspect of trial success.

\item \texttt{patient\_drop}: The number of patients that drop out during the trial. Can be thresholded. Patients drop out from the clinical trial due to lack of efficacy unintended adverse events or other reasons that could result in unanticipated trial outcomes. 

\item \texttt{sites}: The number of total sites during the trial. Can be thresholded. This also indirectly measures the funding capabilities of the sponsors, much like num\_sponsors.

\item \texttt{pvalues}: Whether there exists a primary outcome P-value $<$ 0.05. If there is at least one significant result, we consider this a success. Otherwise, we predict failure.

\item \texttt{update\_more\_recent}: The difference in the date at which the trial was last updated vs its completion date. Can be thresholded. The time gap can provide critical insights into the trials' post-completion process and transparency. Identifying delays is helpful for trial success, as it could be due to data analysis and validation, regulatory review, and publication process. A highly amended trial could indicate success due to the large amount of publications.

\item \texttt{death\_ae, serious\_ae, all\_ae}: Represents the number of deaths, serious adverse events, and total adverse events, respectively. Can be thresholded. The total number of adverse events (AEs), serious adverse events (SAEs), and deaths in a clinical trial can provide important safety information about the investigational treatment. However, the significance of these numbers depends on various factors, including the size and duration of the trial, the nature of the treatment, and the characteristics of the study population. 

\item \texttt{status}: The status of the trial. We say that a trial is not successful if the status is \textit{'Terminated', 'Withdrawn', 'Suspended', 'Withheld', 'No longer available', or 'Temporarily not available'}. However, if it is \textit{'Approved for marketing'}, then we say it is successful. Otherwise, we abstain from predicting either. Having this information incorporated serves to enhance the transparency, regulatory compliance, and ethical conduct of clinical trials. I.e. \textit{"Terminated: The study has stopped early and will not start again. Participants are no longer being examined or treated."} usually occurs when the trial causes significant negative side-effects in several patients.

\item \texttt{amendments}: Represents the number of trial amendments. Can be thresholded.
Clinical trials must follow approved protocols \cite{ICH2016, MHRA2012}, but amendments may be required after regulatory approval to adjust protocols according to new requirements or insights. 
We scraped record histories for each trial from \url{https://clinicaltrials.gov/} and calculated the total number of times a clinical trial has been amended. The number of amendments to a clinical trial protocol can provide some insights into the trial's progress and potential success, but it's not necessarily a direct indication of success or failure. Therefore, we consider the total number of amendments of trial as a weak label. Sometimes, amendments to trial protocols are crucial for adapting to emerging data from ongoing clinical trials, addressing safety concerns, or optimizing study design to yield better outcomes. 

\item \texttt{stock\_price}: Is positive if the sponsor stock price's 5-day moving average has a positive or negative slope. 

\item \texttt{linkage}: Is positive if a trial was found to have any later-stage trials linked to it.

\item \texttt{news\_headlines}: Is positive or negative depending on sentiment from any news headline related to the trial.

\item \texttt{gpt}: Represents GPT3.5-turbo-0125's decisions on PubMed abstracts.

\end{itemize}

\subsection*{Phase-Specific Thresholding}
For Phases 1, 2, and 3, we find specific quantile thresholds from $(0.1, 0.2, ..., 0.9)$ for all LFs that have tunable thresholds, fine-tuned on each respective phase on the TOP training dataset.

To reiterate our final labeling process, we utilize both an unsupervised aggregation approach-- data programming-- and a supervised random forest to obtain our estimated labels and ground our weakly supervised signals on the humanly annotated TOP training data. 

Note that we train all approaches phase-wise. That is, we first select all trials according to Phase 1, train the labeling model, and predict outcome labels for all Phase 1 trials, before moving on to Phase 2. For data programming, we add TOP training labels to all of our other weakly supervised LFs. We duplicate TOP labels 3 times to obtain high agreement and therefore high weight in the matrix completion step.
For our supervised approach, we train a Random Forest model on all other weakly supervised LF outputs to predict the ground truth.

\clearpage
\section*{Ablation on Features for Trial Linkage}
\label{supplementary: Ablation on Features for Trial Linkage}
This section focuses on the ablation study conducted on the trial features for the trial linking algorithm. Figure~\ref{fig:trial_linkage_feat_ablation2} presents the comparison results with TOP and the outcome labels from the trial linkage created using individual features. The features "intervention," "official title," "brief summary," and "eligibility criteria" consistently achieved better performance across all phases. In contrast, "lead sponsor" performed the worst, likely because trial sponsors often change based on funding capacity, even when a trial progresses to the next phase.

Additionally, Table~\ref{tab:feat_comb_trial_linkage} shows the performance of various combinations of trial features across phases. Combining all trial features except "lead sponsor" yielded better performance across all phases. Including "lead sponsor" features reduced the accuracy of the extracted weak outcome labels. However, adding "lead sponsor" improved performance for Phase 1 trials, as it is rare for a trial to change sponsors between Phase 1 and Phase 2. 

\begin{figure}[ht]
    \centering
    \includegraphics[width=.8\linewidth]{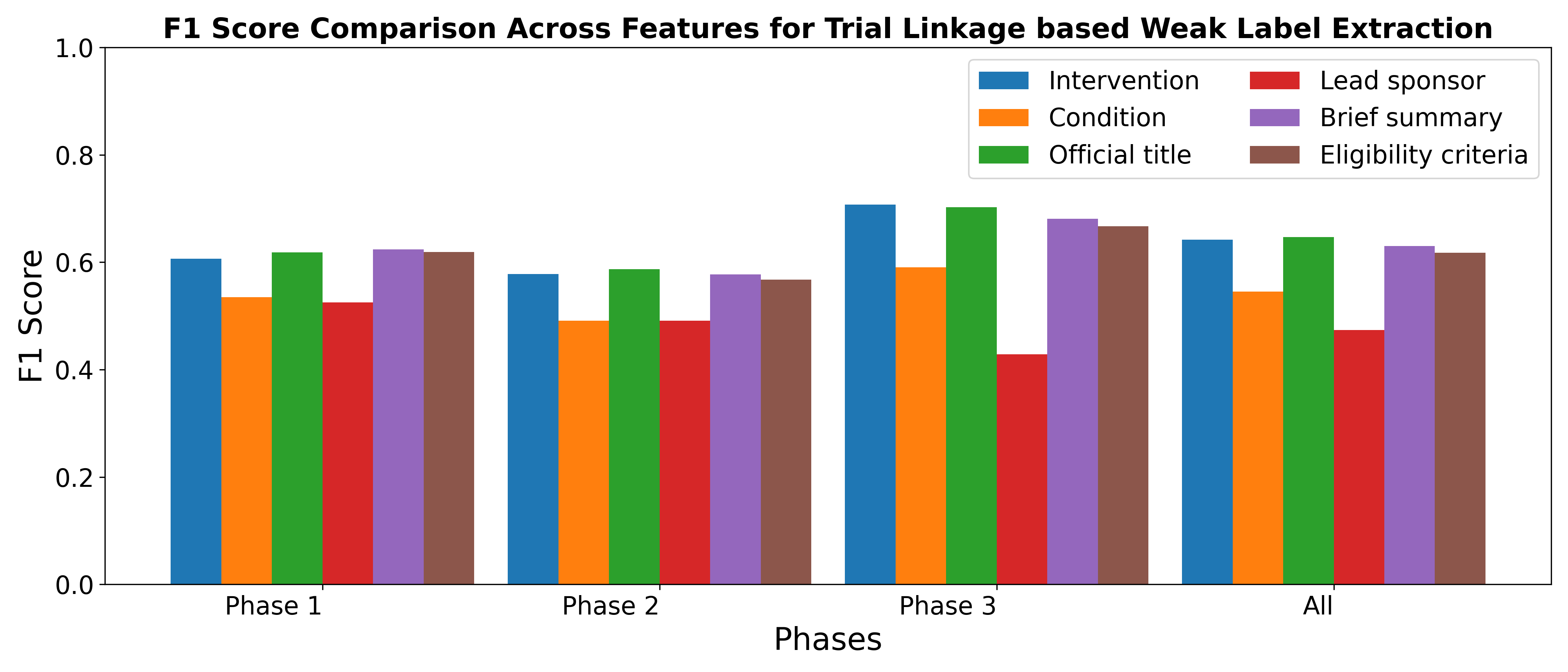}
    \caption{Ablation study on the use of different trial features to predict trial linkages and the agreement of their extracted weak labels with TOP human-annotated data labels using F1 scores.}
    \label{fig:trial_linkage_feat_ablation2}
\end{figure}

\begin{table}[ht]
\centering
\caption{Results of the ablation study on the different combinations of trial features to predict trial linkages and the agreement of their extracted weak labels with TOP human-annotated data labels using F1 scores. \textcolor{red}{$\downarrow$} indicates that adding the feature decreased the performance. \textcolor{green}{$\uparrow$} indicates that adding the feature improved the performance.}
\label{tab:feat_comb_trial_linkage}
\vspace{0.1cm}
\resizebox{\linewidth}{!}{
\begin{tabular}{c|c|c}
Phase & Feature combination & F1
 \\ \midrule
\multirow{5}{*}{Phase 1} & Official title + Intervention & 0.645 \\
& Official title + Intervention + Brief Summary & 0.6426 \textcolor{red}{$\downarrow$}\\
& Official title + Intervention + Brief Summary + Eligibility & 0.6447\textcolor{green}{$\uparrow$}\\
& Official title + Intervention + Brief Summary + Eligibility + Condition & 0.6532 \textcolor{green}{$\uparrow$}\\
& Official title + Intervention + Brief Summary + Eligibility + Condition + Lead Sponsor & \textbf{0.6658} \textcolor{green}{$\uparrow$}\\ \midrule

\multirow{5}{*}{Phase 2} & Official title + Intervention & 0.6027\\
& Official title + Intervention + Brief Summary & 0.6042 \textcolor{green}{$\uparrow$} \\
& Official title + Intervention + Brief Summary + Eligibility & 0.608 \textcolor{green}{$\uparrow$}\\
& Official title + Intervention + Brief Summary + Eligibility + Condition & \textbf{0.6196} \textcolor{green}{$\uparrow$}\\
& Official title + Intervention + Brief Summary + Eligibility + Condition + Lead Sponsor & 0.6189\textcolor{red}{$\downarrow$}\\ \midrule

\multirow{5}{*}{Phase 3} & Official title + Intervention & 0.7323 \\
& Official title + Intervention + Brief Summary & 0.7319\textcolor{red}{$\downarrow$}\\
& Official title + Intervention + Brief Summary + Eligibility & 0.7379\textcolor{green}{$\uparrow$}\\
& Official title + Intervention + Brief Summary + Eligibility + Condition & \textbf{0.7498}\textcolor{green}{$\uparrow$} \\
& Official title + Intervention + Brief Summary + Eligibility + Condition + Lead Sponsor & 0.7384 \textcolor{red}{$\downarrow$}\\ \midrule

\multirow{5}{*}{All} & Official title + Intervention & 0.6686\\
& Official title + Intervention + Brief Summary & 0.6687\textcolor{green}{$\uparrow$}\\
& Official title + Intervention + Brief Summary + Eligibility & 0.6737\textcolor{green}{$\uparrow$}\\
& Official title + Intervention + Brief Summary + Eligibility + Condition & \textbf{0.6842}\textcolor{green}{$\uparrow$} \\
& Official title + Intervention + Brief Summary + Eligibility + Condition + Lead Sponsor & 0.6796\textcolor{red}{$\downarrow$}\\ 
\bottomrule
\end{tabular}
}

\end{table}

\clearpage

\section*{Prompts for LLM Predictions on PubMed Abstracts}
\label{supplementary: Prompts for LLM Predictions}

In this section, we describe the method used to obtain LLM predictions on PubMed abstracts, including the statistical test results and question-answer pairs. An example of such a prompt is shown in Figure~\ref{fig:LLM_prompts}. Additionally, we provide two examples of input abstracts given to the LLM and their resulting outputs, as shown in Figures~\ref{fig:GPT_response example} and \ref{fig:GPT_response example2}.

\begin{figure}[t!]
    \centering
    \includegraphics[width=\linewidth]{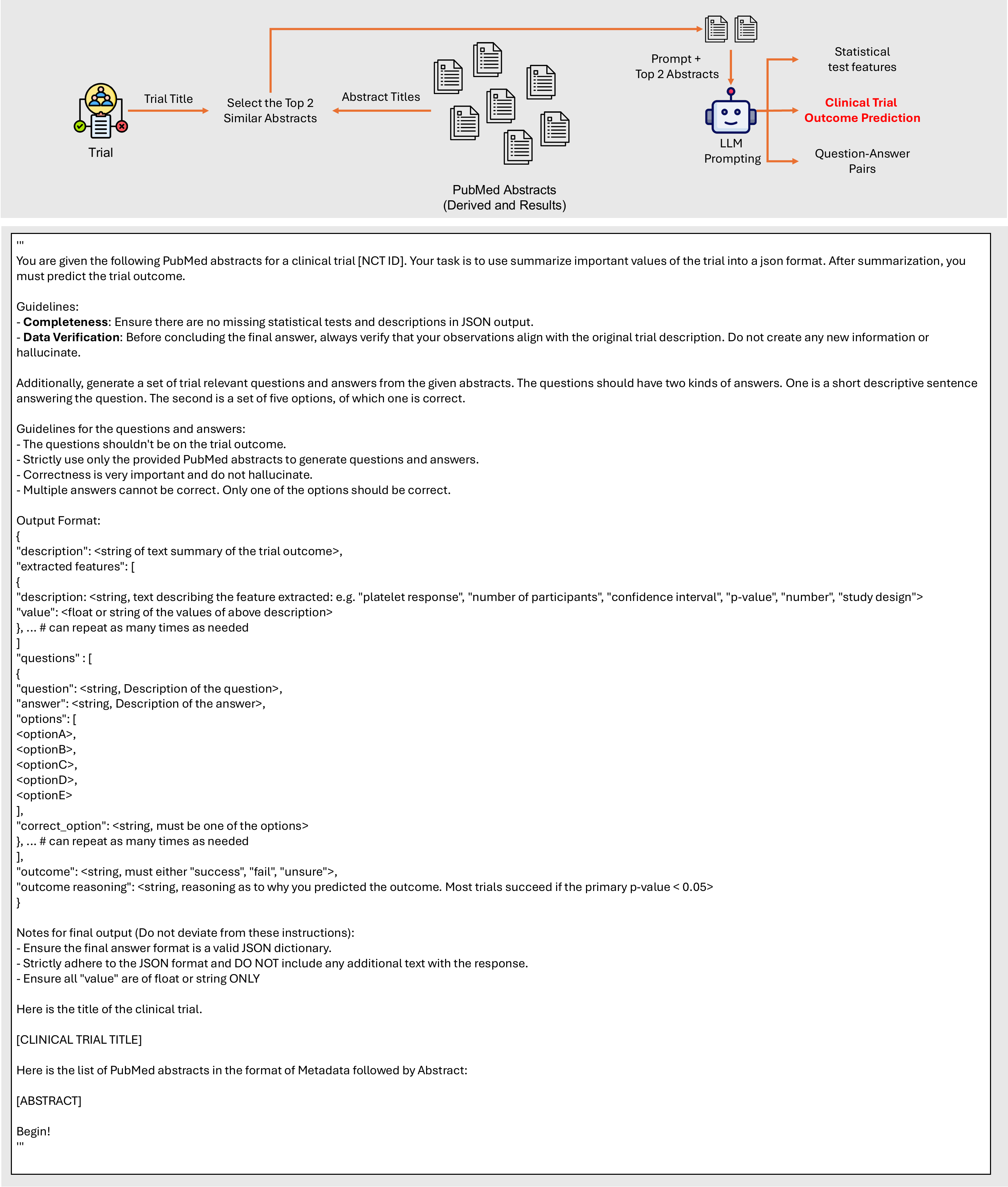}
    \caption{LLM Prompts.}
    \label{fig:LLM_prompts}
\end{figure}

\begin{figure}[ht]
    \centering
    \includegraphics[width=\linewidth]{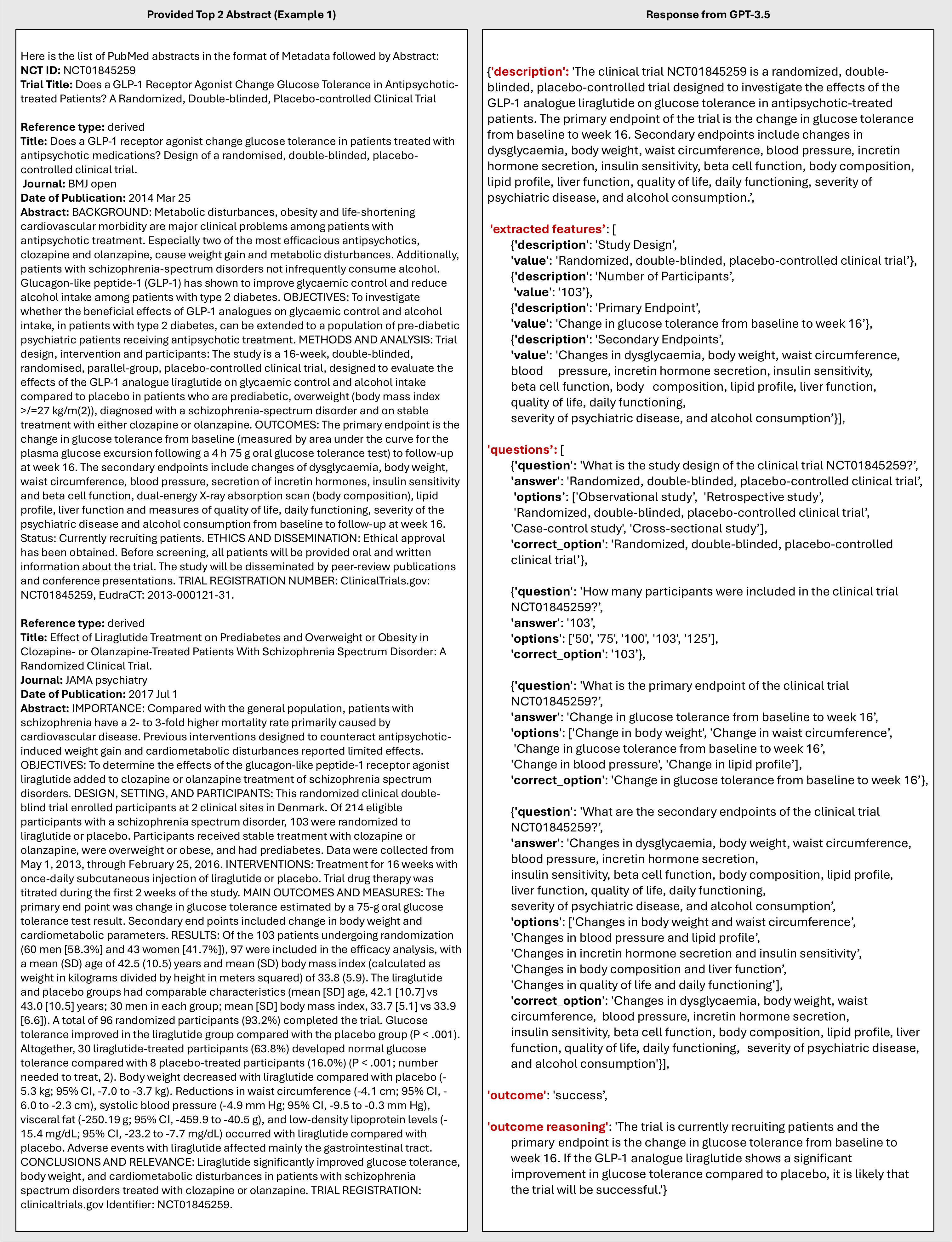}
    \caption{Example response from GPT-3.5.}
    \label{fig:GPT_response example}
\end{figure}

\begin{figure}[ht]
    \centering
    \includegraphics[width=\linewidth]{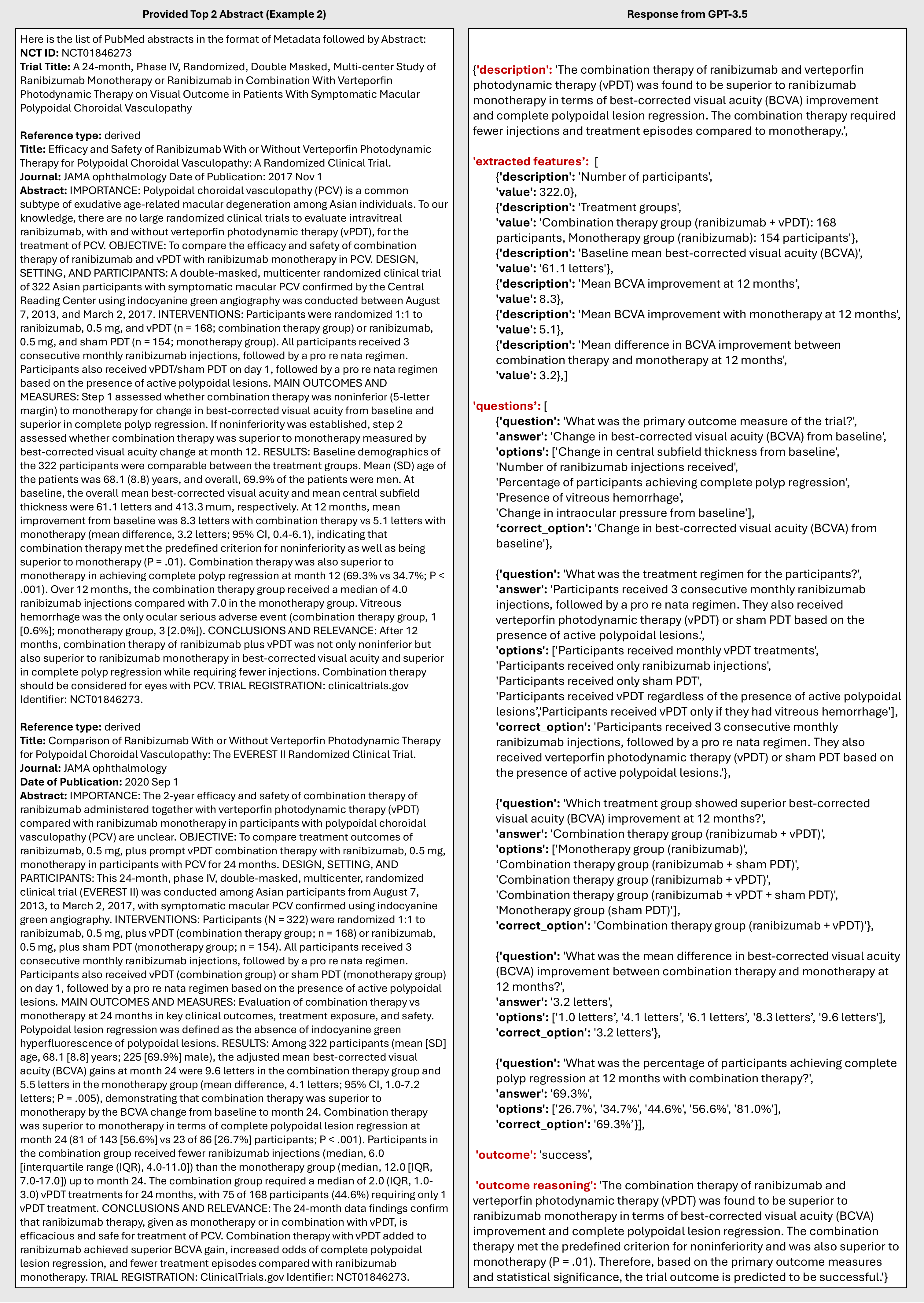}
    \caption{Example response 2 from GPT-3.5.}
    \label{fig:GPT_response example2}
\end{figure}
\clearpage
\newpage
\newpage

\section*{Case studies}
We utilized our random forest labels for the following case studies.

\subsection*{Case study 1} We conducted a case study on clinical trial NCT01213160\footnote{\url{https://clinicaltrials.gov/study/NCT01213160}}. This trial was completed in 2013. Five different weak labels (GPT decision, trial linkage, stock price, sites, and amendment) suggest that the trial was successful. The link to the PubMed article for this trial is \url{https://ncbi.nlm.nih.gov/pmc/articles/PMC5502072/}, which also suggests that \textit{"AZD4547 was well tolerated in Japanese patients, with the best response of stable disease $\geq$ 4 weeks."} Therefore, we believe NCT01213160 was successful, as our CTO label suggests. We believe the TOP label for this trial is incorrect.

\subsection*{Case study 2} Another clinical trial we examined in detail is NCT01111188\footnote{\url{https://clinicaltrials.gov/study/NCT01111188}}. This trial was terminated. Therefore, our label for this trial is Failure (label: 0). However, its TOP label is Success (label: 1), which is unlikely as the trial was not completed. Additionally, the GPT decision for the trial was 0, which means it thinks the PubMed abstract we collected suggests that the trial failed. \textit{"...all patients required dose delays during cycle 2 due to cytopenias, and the study team decided to stop the trial...with the primary toxicity being myelosuppression"}.

\section*{Datasheet (Inspired from Good Practices in AI)}
\paragraph{A.1\;\;\;\;Motivation}
\begin{itemize}
    \item \textbf{For what purpose was the dataset created?}
    
    We created \method\ to democratize clinical trial outcome prediction, which was previously only available to industry-sponsored researchers. Furthermore, we attempt to expand the previous labeling efforts (which were primarily focused on small drug interventions) and predict labels on all trials.
    
    \item \textbf{Who created the dataset (e.g., which team, research group) and on behalf of which entity (e.g., company, institution, organization)?}
    
    The authors of this paper.

    \item \textbf{Who funded the creation of the dataset? If there is an associated grant, please provide the name of the grantor and the grant name and number.}

    This work was supported by NSF award SCH-2205289, SCH-2014438, and IIS-2034479.
    
\end{itemize}

\paragraph{A.2\;\;\;\;Composition}
\begin{itemize}
    \item \textbf{What do the instances that comprise the dataset represent (e.g., documents, photos, people, countries)?}
    
    \method\ contains molecule SMILES strings, eligibility criteria, ICD Codes, drug names, diseases, study status, phase information, and our automatically created labels. Furthermore, we also have additional QA tasks as extracted by GPT 3.5 Turbo 0125, news articles mined for the top 1000 industry sponsors, and stock prices.

    \item \textbf{How many instances are there in total (of each type, if appropriate)?}

    There is a total of 479,761 trials, each with multiple types of predicted labels via random forest and data programming, as well as each phase-optimized threshold. For small-molecule drug interventions, we also have SMILES and ICD Codes. 
    We release all such labels to not limit any downstream use. Furthermore, there are a total of 1,115,017 news articles we extracted and 105,570 trials with corresponding QA pairs.
    
    \item \textbf{Does the dataset contain all possible instances or is it a sample (not necessarily random) of instances from a larger set?}

    We provide all possible trial labels up to the beginning of May 2024. 

    \item \textbf{What data does each instance consist of?}

    \method\ contains molecule eligibility criteria, drug names, diseases, study status, phase information, and our automatically created labels. 
    
    \item \textbf{Is there a label or target associated with each instance?}

    The automatically predicted label is provided for each question.

    \item \textbf{Is any information missing from individual instances? If so, please provide a description, explaining why this information is missing (e.g., because it was unavailable). This does not include intentionally removed information, but might include, e.g., redacted text.}
    
    No.

    \item \textbf{Are relationships between individual instances made explicit (e.g., users' movie ratings, social network links)?}

    No.

    \item \textbf{Are there recommended data splits (e.g., training, development/validation, testing)?}

    Yes, please see Table \ref{tab:humanlabel} for a split of training trials with completion date $<2022$ tested on trials with completion date $\geq 2022$. 

    \item \textbf{Are there any errors, sources of noise, or redundancies in the dataset?}

    Automatically created labels inherently come with an element of noise. However, our high agreement with TOP's human labels (up to 0.91 F1), implies that our labels are of high quality.


    \item \textbf{Is the dataset self-contained, or does it link to or otherwise rely on external resources (e.g., websites, tweets, other datasets)?}

    \method\ depends on multiple open source datasets.
    \begin{enumerate}
        \item CTTI: \url{https://ctti-clinicaltrials.org/} 
        \item DrugBank \url{https://drugbank.com/}
        \item PubMed \url{https://pubmed.ncbi.nlm.nih.gov/}
    \end{enumerate}


    \item \textbf{Does the dataset contain data that might be considered confidential (e.g., data that is protected by legal privilege or by doctor-patient confidentiality, data that includes the content of individuals' non-public communications)?}

    No. We obtained all data sources via open-source methods.

    \item \textbf{Does the dataset contain data that, if viewed directly, might be offensive, insulting, threatening, or might otherwise cause anxiety?}

    No.

    \item \textbf{Does the dataset relate to people?}

    Yes.

    \item \textbf{Does the dataset identify any subpopulations (e.g., by age, gender)?}
    
    Yes, but only in the eligibility criteria of the trials, which are public.

    \item \textbf{Does the dataset contain data that might be considered sensitive in any way (e.g., data that reveals race or ethnic origins, sexual orientations, religious beliefs, political opinions or union memberships, or locations; financial or health data; biometric or genetic data; forms of government identification, such as social security numbers; criminal history)?}

    No. There is no reference to individuals.
    
\end{itemize}

\paragraph{A.3\;\;\;\;Collection process}
\begin{itemize}
    \item \textbf{How was the data associated with each instance acquired?}

    We automatically mined each LLM prediction, trial linkage, news headline, and stock price as an overview in Section \method\ Overview.

    \item \textbf{What mechanisms or procedures were used to collect the data (e.g., hardware apparatuses or sensors, manual human curation, software programs, software APIs)?}
    
    We mainly used Google Sheets and Python to collect, process, and label the data.
    In addition, we used OpenAI's ChatGPT (GPT-3.5-turbo 0125) to generate QA and GPT outcome predictions.

    \item \textbf{If the dataset is a sample from a larger set, what was the sampling strategy (e.g., deterministic, probabilistic with specific sampling probabilities)?}

    Some data splitting was done according to previous data splits from TOP \url{https://github.com/futianfan/clinical-trial-outcome-prediction}. Additionally, we also split the data according to the following dates: $(-\infty,2018, 2020, \infty)$
        
    \item \textbf{Who was involved in the data collection process (e.g., students, crowd workers, contractors) and how were they compensated (e.g., how much were crowd workers paid)?}

    The authors of the paper collected and processed the data.

    \item \textbf{Over what timeframe was the data collected?}

    We collected the data between December 2023 and April 2024.

    \item \textbf{Were any ethical review processes conducted (e.g., by an institutional review board)?}

    N/A.

    \item \textbf{Does the dataset relate to people?}

    Yes.

    \item \textbf{Did you collect the data from the individuals in question directly, or obtain it via third parties or other sources (e.g., websites)?}

    N/A.

    \item \textbf{Were the individuals in question notified about the data collection?}

    N/A.

    \item \textbf{Did the individuals in question consent to the collection and use of their data?}

    N/A.

    \item \textbf{If consent was obtained, were the consenting individuals provided with a mechanism to revoke their consent in the future or for certain uses?}

    N/A.

    \item \textbf{Has an analysis of the potential impact of the dataset and its use on data subjects (e.g., a data protection impact analysis) been conducted?}

    The dataset does not have individual-specific information.
\end{itemize}

\paragraph{A.4\;\;\;\;Preprocessing/cleaning/labeling}
\begin{itemize}
    \item \textbf{Was any preprocessing/cleaning/labeling of the data done (e.g., discretization or bucketing, tokenization, part-of-speech tagging, SIFT feature extraction, removal of instances, processing of missing values)?}

    N/A.

    \item \textbf{Was the ``raw'' data saved in addition to the preprocess/cleaned/labeled data (e.g., to support unanticipated future uses)?}

    N/A.

    \item \textbf{Is the software that was used to preprocess/clean/label the data available?}

    Preprocessing, cleaning, and labeling are done via Google Sheets and Python.   
\end{itemize}

\paragraph{A.5\;\;\;\;Uses}
\begin{itemize}
    \item \textbf{Has the dataset been used for any tasks already?}

    No.

    \item \textbf{Is there a repository that links to any or all papers or systems that use the dataset?}

    No.

    \item \textbf{What (other) tasks could the dataset be used for?}

    Our dataset is designed to promote research primarily in clinical trial outcome prediction. The dataset can also be used for stock price/trend prediction, question answering, etc. 

    \item \textbf{Is there anything about the composition of the dataset or the way it was collected and preprocessed/cleaned/labeled that might impact future uses?}

    N/A.

    \item \textbf{Are there tasks for which the dataset should not be used?}

    N/A.
\end{itemize}

\paragraph{A.6\;\;\;\;Distribution}
\begin{itemize}
    \item \textbf{Will the dataset be distributed to third parties outside of the entity (e.g., company, institution, organization) on behalf of which the dataset was created?}

    No.

    \item \textbf{How will the dataset be distributed?}

    Since clinical trial data is frequently updated, we provide the code for generating our CTO dataset at \url{https://github.com/chufangao/CTOD}. The current version of the dataset can be accessed at \url{https://zenodo.org/doi/10.5281/zenodo.11535960}.
    

    \item \textbf{Will the dataset be distributed under a copyright or other intellectual property (IP) license, and/or under applicable terms of use (ToU)?}

    The dataset is released under MIT License.

    \item \textbf{Have any third parties imposed IP-based or other restrictions on the data associated with the instances?}

    No.

    \item \textbf{Do any export controls or other regulatory restrictions apply to the dataset or to individual instances?}

    No.
\end{itemize}

\paragraph{A.7\;\;\;\;Maintenance}
\begin{itemize}
    \item \textbf{Who will be supporting/hosting/maintaining the dataset?}

    The authors of this paper.

    \item \textbf{How can the owner/curator/manager of the dataset be contacted(e.g., email address)?}
    
    Contact the corresponding authors (\texttt{chufan2@illinois.edu} \& \texttt{jp65@illinois.edu} \& \texttt{trishad2@illinois.edu} \& \texttt{jimeng@illinois.edu}).

    \item \textbf{Is there an erratum?}

    No.

    \item \textbf{Will the dataset be updated (e.g., to correct labeling errors, add new instances, delete instances)?}

    If any corrections are needed, we plan to upload an updated version of the dataset along with detailed explanations of the changes. 

    \item \textbf{If the dataset relates to people, are there applicable limits on the retention of the data associated with the instances (e.g., were the individuals in question told that their data would be retained for a fixed period of time and then deleted)?}

    N/A

    \item \textbf{Will older versions of the dataset continue to be supported/hosted/maintained?}

    Primarily, we aim to keep only the latest version of the dataset. However, in specific cases like major updates to the dataset or the necessity to validate previous research with older versions, we will exceptionally retain past versions of the dataset for up to one year.

    \item \textbf{If others want to extend/augment/build on/contribute to the dataset, is there a mechanism for them to do so?}

    Contact the authors of this paper or raise a github issue.
\end{itemize}
\end{document}